\documentclass[pdflatex,sn-mathphys-num]{sn-jnl}

\usepackage{graphicx}%
\usepackage{multirow}%
\usepackage{amsmath,amssymb,amsfonts}%
\usepackage{amsthm}%
\usepackage{mathrsfs}%
\usepackage[title]{appendix}%
\usepackage{xcolor}%
\usepackage{textcomp}%
\usepackage{manyfoot}%
\usepackage{booktabs}%
\usepackage{algorithm}%
\usepackage{algorithmicx}%
\usepackage{algpseudocode}%
\usepackage{listings}%
\usepackage{fullpage}%
\usepackage{multirow}%
\usepackage{makecell}%
\usepackage{caption}

\theoremstyle{thmstyleone}%
\theoremstyle{thmstyletwo}%

\theoremstyle{thmstylethree}%

\begin{document}

\title[Article Title]{Analogical Reasoning as a Doctor: A Foundation Model for Gastrointestinal Endoscopy Diagnosis}


\author[1]{\fnm{Peixi} \sur{Peng}}\email{xiandyxi@sjtu.edu.cn}
\equalcont{These authors contributed equally to this work.}

\author[1]{\fnm{Housheng} \sur{Xie}}\email{housheng.xie@sjtu.edu.cn}
\equalcont{These authors contributed equally to this work.}

\author[2]{\fnm{Yanling} \sur{Wei}}\email{lingzi016@tmmu.edu.cn}
\equalcont{These authors contributed equally to this work.}

\author[2]{\fnm{Guangcong} \sur{Ruan}}\email{ruanguangcong@tmmu.edu.cn}
\equalcont{These authors contributed equally to this work.}

\author[1]{\fnm{Xiaoyang} \sur{Zou}}\email{xiaoyang.zou@sjtu.edu.cn}

\author[3]{\fnm{Qian} \sur{Cao}}\email{caoq@zju.edu.cn}

\author[2]{\fnm{Yongjian} \sur{Nian}}\email{yongjian\_nian@163.com}

\author*[1]{\fnm{Guoyan} \sur{Zheng}}\email{guoyan.zheng@sjtu.edu.cn}

\affil*[1]{\orgdiv{Institute of Medical Robotics}, \orgname{School of Biomedical Engineering}, \orgaddress{\street{Shanghai Jiao Tong University}, \city{Shanghai}, \postcode{200240}, \country{China}}}

\affil[2]{\orgdiv{Department of Gastroenterology}, \orgname{Daping Hospital}, \orgaddress{\street{Army Medical University (Third Military Medical University)}, \city{Chongqing}, \postcode{400042}, \country{China}}}

\affil[3]{\orgdiv{Department of Gastroenterology}, \orgname{Sir Run Run Shaw Hospital}, \orgaddress{\street{Zhejiang University School of Medicine}, \city{Hangzhou}, \postcode{310016}, \country{China}}}


\abstract{Gastrointestinal diseases represent a growing global health burden, for which endoscopy serves as a primary tool for early diagnosis. However, routine endoscopic image diagnosis still suffers from missed lesions and suboptimal efficiency. While Artificial intelligence (AI)-assisted diagnosis has shown great promise, existing AI models remain limited in generalizability, adaptability, robustness, and scalability, primarily due to medical data scarcity, domain shift, and annotation heterogeneity. To address these limitations, we developed a relevance-knowledge acquisition and transfer network (RATNet), which is a foundation model for gastrointestinal endoscopy imaging based on analogical reasoning. RATNet acquires and transfers knowledge from heterogeneous expert annotations across five gastrointestinal endoscopy datasets through a cyclic pre-training strategy. Its architecture comprises an encoder, a RAT module, a projector, and a multi-task head, supporting fine-tuning, linear probing, and zero-shot transfer.
Comprehensive evaluations demonstrate that RATNet outperforms existing foundation models (e.g., GastroNet, GastroVision) across six clinical scenarios: it accurately diagnoses common gastrointestinal diseases, learns from few samples for rare diseases, achieves zero-shot transfer to new medical sites, handles long-tailed disease distributions, adapts to novel diseases, and safeguards patient privacy via federated learning. The superior performance stems from its analogical reasoning mechanism, which mimics clinicians’ cognition by matching the image-derived posterior knowledge to a learned prior knowledge base and transferring the acquired relative knowledge to guide diagnosis, thereby improving generalization and bias resistance.
RATNet is an open and cost-effective model that supports automatic integration of heterogeneous annotations without manual label unification, significantly reducing data acquisition costs. By fostering collaboration and open access, RATNet is poised to become a cornerstone for intelligent gastrointestinal diagnosis, accelerating AI adoption in resource-limited settings.}




\maketitle

\section{Introduction}\label{sec1}

Gastrointestinal (GI) diseases are increasingly prevalent worldwide~\cite{wang2023global,arnold2020global}, with digestive tract cancers long occupying a prominent position in the global cancer burden~\cite{sung2021global}. The prognosis for these cancers is generally poor, particularly at advanced stages, where survival rates drop significantly~\cite{correa2013gastric}, underscoring the critical importance of early screening. Endoscopy serves as the primary diagnostic tool for detecting early pathological changes~\cite{tang2021advances,martins2023endoscopic}. However, current practice relying on manual endoscopic image diagnosis faces challenges such as missed diagnoses and limited efficiency in image review. Artificial intelligence (AI)-assisted diagnosis offers a promising approach to address these limitations~\cite{tham2025artificial,xu2025artificial,mushtaq2026ai}.

Foundation models, which are large-scale deep learning architectures pre-trained on extensive and diverse datasets, are reshaping the landscape of artificial intelligence in healthcare. By training on vast and diverse data, these models learn universal feature representations and can be fine-tuned for specific tasks with minimal additional data~\cite{shi2024survey,zhang2024challenges}. Leveraging their comprehensive pre-training, they demonstrate diagnostic accuracy for anomalies that surpasses traditional methods and even human experts, while also exhibiting strong generalization to clinical scenarios beyond their initial training scope~\cite{ma2025fully}. This enables a wide range of applications from radiology~\cite{wu2025towards} and pathology~\cite{wang2024pathology} to endoscopy~\cite{boers2024foundation}. Consequently, foundation models hold promise for democratizing expert-level gastrointestinal diagnostic capabilities, bridging the gap in endoscopic expertise for resource-limited or remote areas.

Although foundation models offer a promising paradigm for developing generalizable and annotation-efficient solutions for gastrointestinal visual tasks, their success heavily depends on access to large and diverse datasets. However, in the field of gastrointestinal diseases, large-scale datasets are relatively scarce, and many foundation models~\cite{he2025foundational,zhang2024learning,dermyer2025endodino} have not been made fully publicly available, thereby hindering further progress within the research community. Fully open public datasets~\cite{wang2020improved,polat2023improving,pogorelov2017kvasir,borgli2020hyperkvasir} are often small and suffer from inconsistent annotations in terms of disease coverage (i.e., the label heterogeneity problem). This makes it challenging to train powerful and robust foundation models by aggregating numerous small public datasets.
Furthermore, endoscopic imaging involves considerable real-world variability~\cite{devkota2025federated} due to diverse equipment, imaging protocols, and patient populations. This variability leads to significant differences in the data distribution across datasets, a problem commonly referred to as domain shift~\cite{kondrateva2021domain}. Existing foundation models lack the ability to extract domain-related knowledge from multiple distinct domains. Consequently, when encountering an unseen target domain, they cannot effectively transfer knowledge from source domains based on domain similarity~\cite{ayana2024multistage}, leading to performance degradation. For instance, the GastroNet-5M dataset~\cite{boers2024foundation} contains endoscopic images recorded using three types of endoscope equipment: Olympus, Pentax, and Fujifilm. The foundation model GastroNet~\cite{boers2024foundation}, pre-trained on GastroNet-5M, generally fails to distinguish between these three domains. When this model performs inference on the PolypGen dataset~\cite{ali2023multi}, which consists solely of images captured with Olympus equipment, it tends to exhibit a bias toward domains with more abundant pre-training data, rather than prioritizing knowledge from the Olympus domain, which shares the same equipment type as PolypGen~\cite{ali2023multi}. This limitation undermines the model's reliability.

\begin{figure}[t]
\centering
\includegraphics[width=\textwidth]{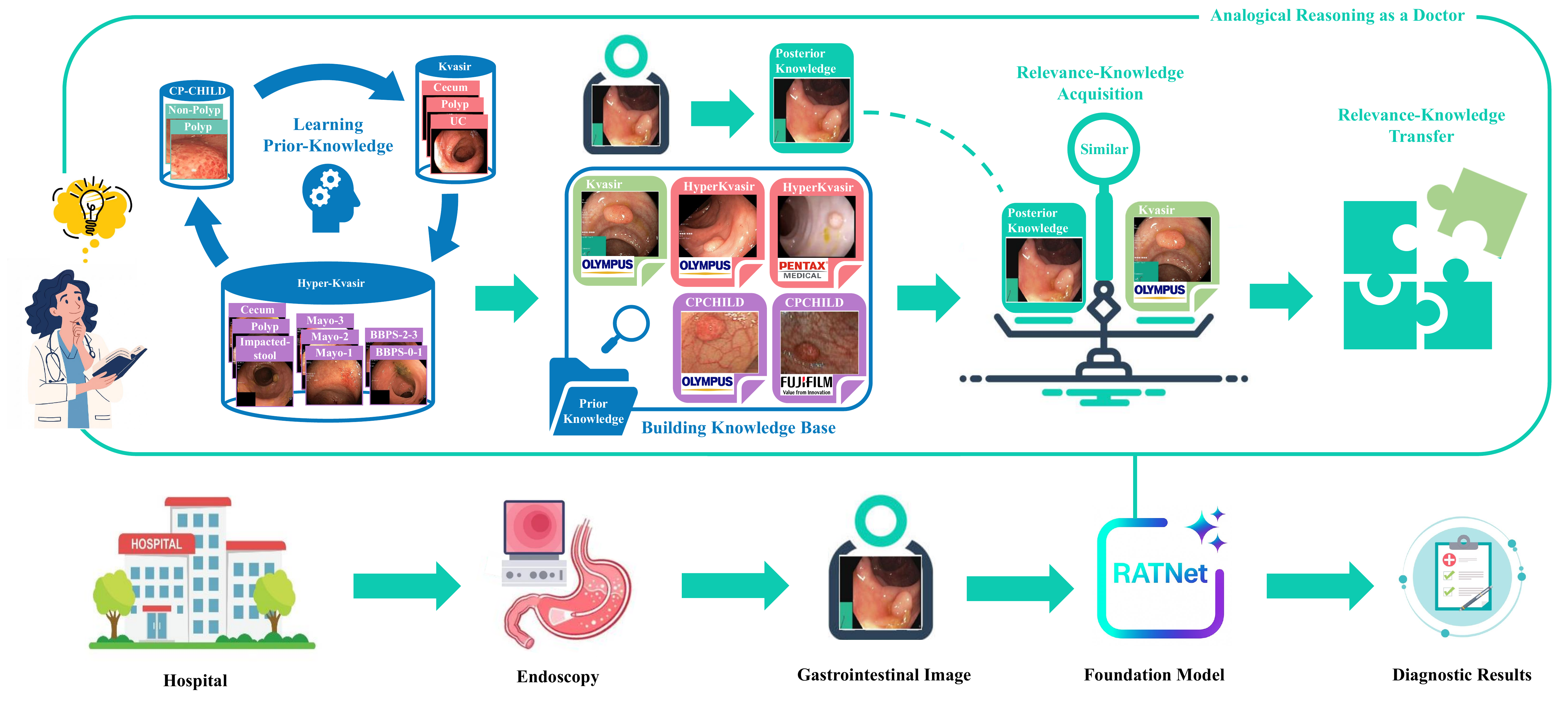}
\caption{Workflow of the base model RATNet. This model utilizes gastrointestinal images obtained from medical endoscopy to provide diagnostic assistance. RATNet simulates the analogical reasoning process of physicians by first learning from multiple datasets to construct a prior knowledge base. It then estimates posterior knowledge from the current gastrointestinal images. By comparing and acquiring relevant knowledge from the prior knowledge base, the model transfers this knowledge to the domain of the posterior knowledge, enabling effective reasoning for diagnostic outcomes.}\label{fig:analogy}
\end{figure}

Research on human cognition offers a promising avenue for addressing the aforementioned limitations, primarily through the application of analogical reasoning. Analogical reasoning~\cite{gentner2017analogical}, which is the ability to perceive and use relational similarity between two situations or events, is a fundamental aspect of human cognition and serves as a core process in scientific discovery, problem-solving, categorization, and decision-making. This mechanism is an important element of various cognitive abilities including inference and abstraction~\cite{gust2008analogical}, enables humans to identify abstract patterns in seemingly disparate environments, transfers knowledge from familiar domains to unfamiliar ones, and generates creative solutions for novel problems~\cite{gentner1997reasoning}. In medical contexts, analogical reasoning mirrors cognitive processes clinicians used for diagnosis. Durning et al.~\cite{guallart2014analogical} highlight its role in two key stages: pattern recognition during data gathering, where doctors identify relational similarities in symptoms, and hypothesis generation, drawing inferences from analogous cases.
Inspired by this, as shown in Figure~\ref{fig:analogy}, we propose a foundation model that mimics a clinician's analogical reasoning process, termed the Relevance-knowledge Acquisition and Transfer Network (RATNet). During training, RATNet iteratively acquires prior knowledge from various datasets. During inference, it estimates posterior knowledge from the input sample and compares it with various prior knowledge from multiple sources to identify relevant associations. This facilitates the transfer of knowledge from the source to the target domain, enabling effective reasoning in unseen domains.
This foundation model can be trained by aggregating numerous public and private datasets without requiring label unification, facilitating broader collaboration within the research community. Furthermore, RATNet can be iteratively refined and enhanced by acquiring prior knowledge from new datasets, enabling the model to adapt to the evolving diagnostic needs. This continuously iterative process of analogical reasoning facilitates robust generalization and predictive inference, while mitigating challenges such as data scarcity and domain-specific overfitting. Such advancements contribute to the progress of AI-driven healthcare, ultimately leading to improved patient outcomes.

RATNet is pre-trained by cyclically acquiring and transferring knowledge embedded within heterogeneous expert annotations from five gastrointestinal endoscopy datasets. It comprises four key components: an encoder, a RAT module, a projector, and a multi-task classifier. These components facilitate flexible transition to clinical applications through three distinct paradigms: full fine-tuning, linear probing, or zero-shot transfer. Full fine-tuning entails updating all network parameters to maximize adaptability, allowing the model to capture domain-specific features and achieve peak performance when sufficient data is available. In contrast, linear probing (or head-only training) keeps the pre-trained backbone entirely frozen and only trains a newly added linear classifier. This computationally efficient method serves as a benchmark to evaluate the quality of the learned representations. Finally, zero-shot transfer assesses the model’s inherent generalization by directly deploying the pre-trained network to diagnose novel conditions in unseen datasets without any task-specific retraining or prior exposure to the samples.

To comprehensively evaluate the capability of RATNet in gastrointestinal disease diagnosis, we conducted extensive benchmarking across ten independent datasets. Five datasets were specifically allocated for in-domain validation, namely CP-CHILD~\cite{wang2020improved}, Kvasir~\cite{pogorelov2017kvasir}, HyperKvasir~\cite{borgli2020hyperkvasir}, LIMUC~\cite{polat2023improving}, and Daping. Of these, Daping is a private dataset, while the others are publicly available. The remaining six datasets were used for cross-institutional out-of-domain generalization assessment, including Colonoscopic~\cite{mesejo2016computer}, PolypGen~\cite{ali2023multi}, GastroVision~\cite{jha2023gastrovision}, Kvasir-Capsule~\cite{smedsrud2021kvasir}, and Shaoyifu. Among these, only Shaoyifu is a private dataset, while the others are publicly accessible (Extended Data Table 1). Please note that Kvasir, HyperKvasir, and Kvasir-Capsule are three distinct datasets, each with unique images and corresponding labels, and they are independent of one another. In-domain validation was performed on held-out test sets from datasets involved in the pretraining phase, primarily examining model performance under similar data distributions. Out-of-domain testing, targeted completely unseen datasets from different medical institutions, encompassing diverse populations and imaging protocols, aiming to simulate distribution shifts commonly encountered in clinical practice and thereby reveal the model's robustness in real-world deployment. Through these diverse data resources, we focused on evaluating RATNet's performance across six critical downstream tasks: classification of common gastrointestinal diseases, rare disease recognition under few-shot conditions, handling of long-tailed imbalanced distributions, diagnostic environment transfer without fine-tuning, zero-shot response to emerging diseases, and privacy-preserving distributed learning based on federated pretraining. Experimental results demonstrate that RATNet significantly outperforms existing foundation models across multiple metrics including generalization capability, adaptability, robustness, and scalability, exhibiting stronger potential for clinical applications.

\begin{figure}[!htbp]
\centering
\includegraphics[width=\textwidth]{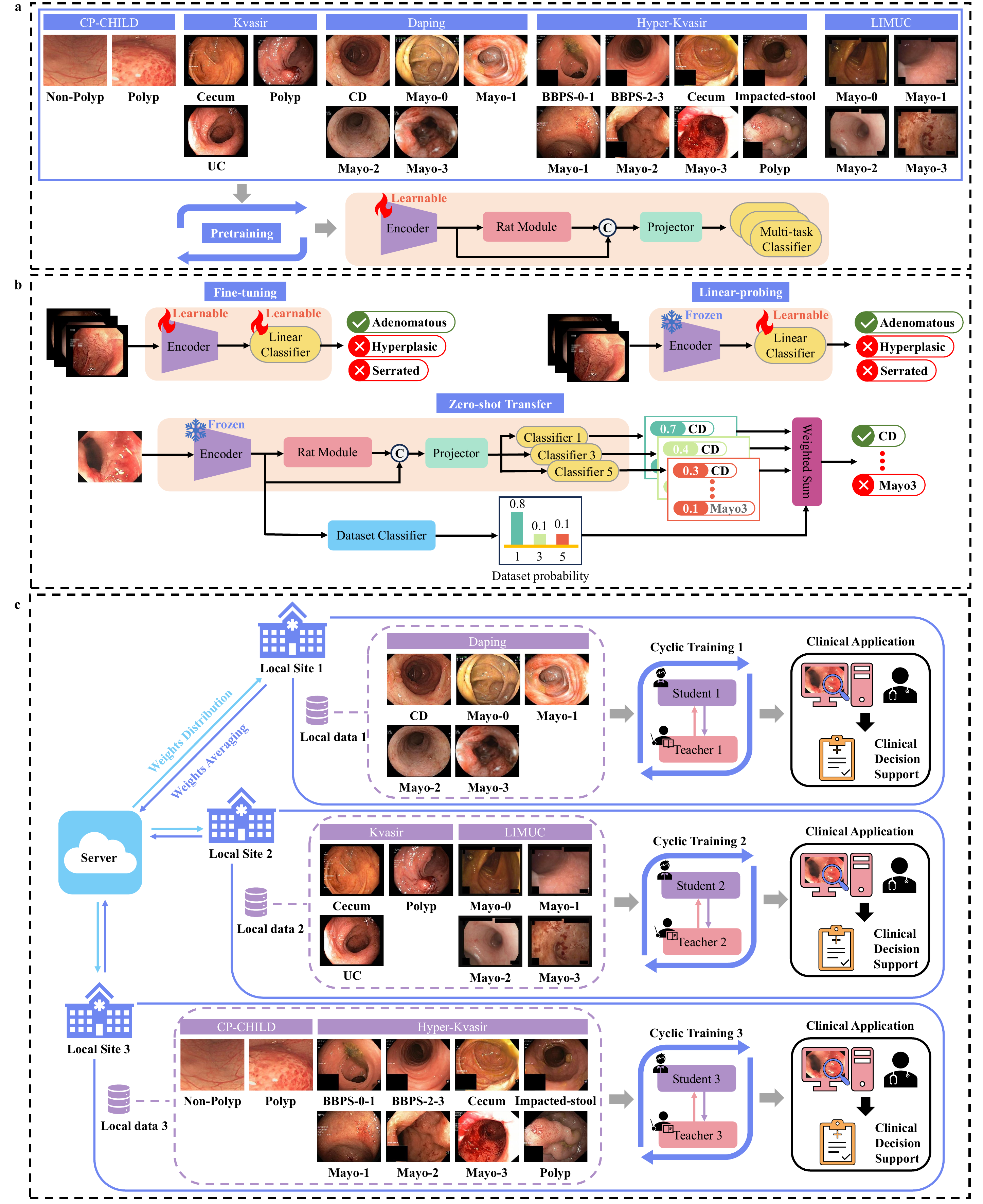}
\caption{Relevance-knowledge acquisition and transfer network (RATNet) is a foundation model for gastrointestinal disease diagnosis. (\textbf{a}) RATNet is pre-trained cyclically using heterogeneous expert annotations from five gastrointestinal endoscopy datasets: CP-CHILD~\cite{wang2020improved}, LIMUC~\cite{polat2023improving}, HyperKvasir~\cite{borgli2020hyperkvasir}, Daping, and Kvasir~\cite{pogorelov2017kvasir}. (\textbf{b}) The model comprises four pre-trained components: encoder, RAT module, projector, and multi-task head, adaptable via fine-tuning, linear probing, or zero-shot transfer. Fine-tuning trains the entire model with a new classifier on target data. Linear probing trains only a new classifier on pre-extracted embeddings. Zero-shot transfer aggregates predictions from all classification heads using domain similarity weights, requiring no additional training. (\textbf{c}) In federated learning, local RATNet models train at each site using cyclic training for student models and EMA for teacher models. After each round, student weights are averaged at a central server to create a global model, which is redistributed for iterative improvement.}\label{fig:overview}
\end{figure}

\section{Results}\label{sec2}

\subsection{Diagnose common gastrointestinal diseases}\label{subsec21}

\begin{figure}[htbp]
\centering
\includegraphics[width=\textwidth]{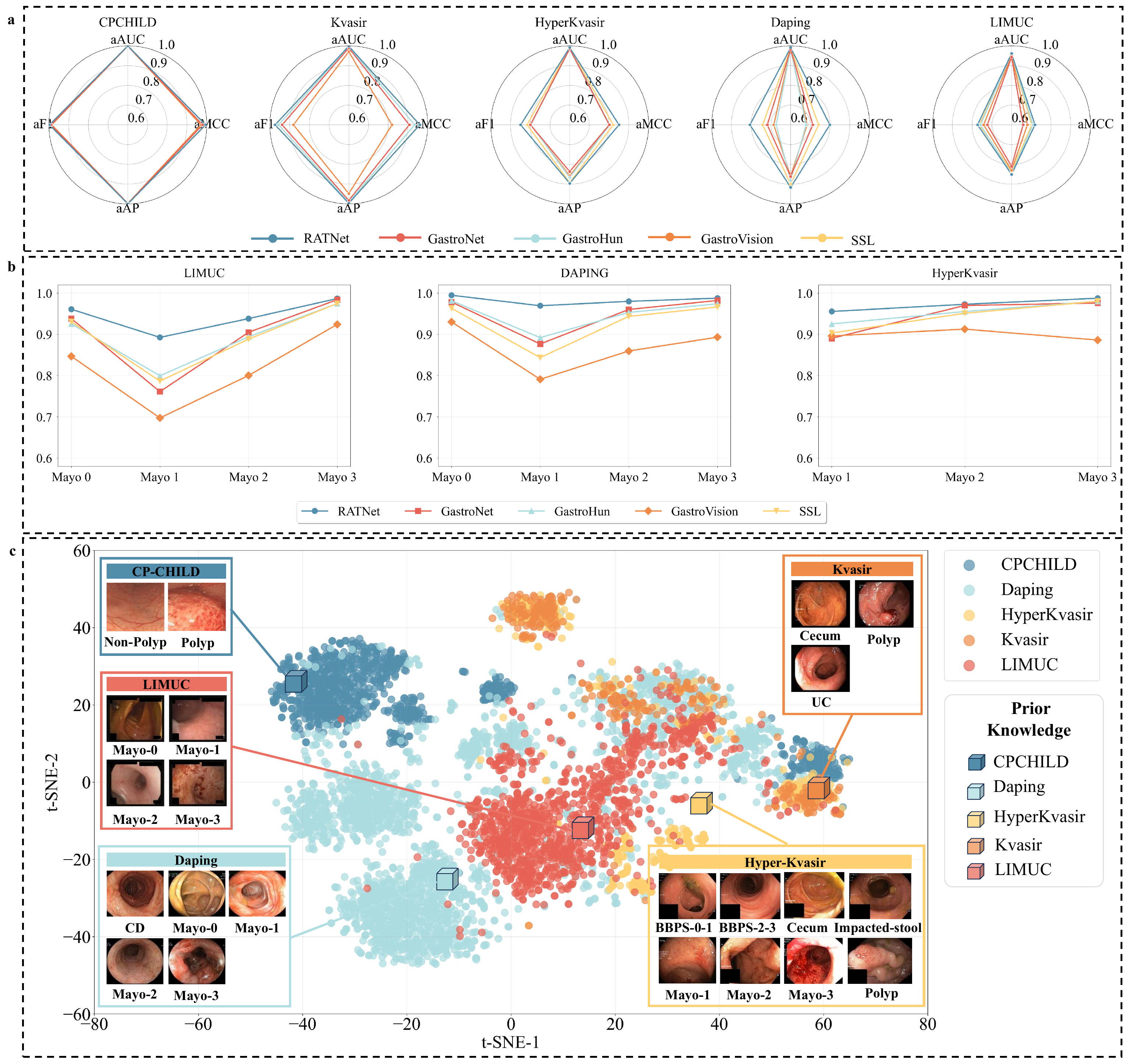}
\caption{Diagnostic performance for common gastrointestinal diseases on internal test sets. \textbf{a.} Comparison of five large pre-trained models in diagnosing gastrointestinal diseases across five internal test sets. The radar charts show the average scores for metrics such as AUC, MCC, F1, and AP across all diseases within each dataset, highlighting the overall diagnostic performance of each model. \textbf{b.} Comparison of the domain generalization ability of five large pre-trained models. We compared the AUC scores of the five models for Mayo endoscopic scoring across three internal test sets, including LIMUC~\cite{polat2023improving}, Daping, and HyperKvasir~\cite{borgli2020hyperkvasir}, to assess their generalization ability. \textbf{c.} Two-dimensional t-distributed Stochastic Neighbor Embedding (t-SNE)~\cite{van2008visualizing} visualization of prior and posterior knowledge acquired by RATNet from the pre-training dataset. This visualization was used to evaluate whether the model can correctly distinguish posterior knowledge from different tasks and successfully extract a representative prior knowledge base.}\label{fig:diagnosis}
\end{figure}

We conducted an internal evaluation of the RATNet model using test data from four public gastrointestinal endoscopy image datasets including CP-CHILD~\cite{wang2020improved}, Kvasir~\cite{pogorelov2017kvasir}, HyperKvasir~\cite{borgli2020hyperkvasir}, LIMUC~\cite{polat2023improving}, and one private dataset termed Daping. Additionally, we compared its performance on these datasets against several large pre-trained models, including GastroNet~\cite{boers2024foundation}, SSL~\cite{bravo2025self}, GastroHUN~\cite{bravo2025gastrohun}, and GastroVision~\cite{jha2023gastrovision}. As illustrated in Figure \ref{fig:diagnosis}(a), RATNet achieved AUC scores above 90\% across all five datasets, demonstrating its generalizability. Furthermore, we compared RATNet with several large vision-language models on the test set of the ColonINST dataset~\cite{ji2026frontiers}. As shown in Extended Data Table 2, RATNet outperformed large vision-language models trained on large-scale datasets, highlighting its ability to iteratively accumulate and leverage knowledge from diverse datasets, thereby improving data utilization efficiency.


To evaluate the quality of the model features, we performed linear probing on five models across the HyperKvasir, LIMUC, and Daping datasets using the Mayo endoscopic score. These three datasets were selected for evaluation since they are the only ones that include the Mayo endoscopic score. The Mayo endoscopic score was chosen for this assessment because it is a widely recognized and clinically relevant metric for determining the severity of colonic mucosal inflammation in patients with inflammatory bowel disease (IBD). Its widespread use among clinicians highlights its importance in both clinical decision-making and model evaluation~\cite{xu2022mayo}.
Furthermore, given that the score assesses the severity of colonic mucosal inflammation, the scoring system requires models to identify subtle differences in image features~\cite{xu2022patch} and accurately distinguish between different levels of inflammation. Using the Mayo endoscopic score in linear probing provides a comprehensive evaluation of the model's ability to extract and represent visual features across different datasets. As shown in Figure \ref{fig:diagnosis}(b), our RATNet model maintained high AUC values (all exceeding 89\%) for Mayo endoscopic score estimation across different datasets, outperforming all other models. This demonstrates RATNet's superior ability in visual feature extraction.

The RAT module acquires prior knowledge from previous tasks to construct a prior knowledge base, and then transfers prior knowledge associated with posterior knowledge to the current task. To demonstrate the effectiveness of our proposed RAT module, we employed t-SNE~\cite{van2008visualizing} to project and visualize the prior and posterior knowledge features of the RATNet model across five different tasks from the pre-training dataset.  Specifically, we extracted posterior knowledge features from the test sets of CP-CHILD, Kvasir, HyperKvasir, LIMUC, and Daping using RATNet, and projected them together with the pre-trained prior knowledge base into a shared two-dimensional space. As shown in Figure \ref{fig:diagnosis}(c), the model effectively learns distinct prior knowledge from the five different tasks in the pre-training dataset, with no overlap between them. Moreover, the prior and posterior knowledge of the same task exhibit greater similarity. Additionally, the fact that the posterior knowledge features of different tasks cluster in separate regions indicates that the RAT module, by repeatedly acquiring and transferring relevance-knowledge, acquires a strong ability to distinguish posterior knowledge features across tasks.

\subsection{Learn rare conditions from a few samples}\label{subsec22}

Precise detection of uncommon gastrointestinal pathologies is vital for optimizing therapeutic interventions and patient prognosis. However, their infrequent presentation creates persistent obstacles for clinicians. The paucity of labeled instances markedly impedes the ability of standard deep learning systems to develop reliable feature representations.
A central metric for gauging the clinical relevance of vision foundation models lies in their robustness when adapting to low-prevalence entities with minimal supervision. To probe this capacity in RATNet, we conducted few-shot evaluations using a linear probing protocol, benchmarking against established pretrained architectures.

\begin{figure}[htbp!]
\centering
\includegraphics[width=\textwidth]{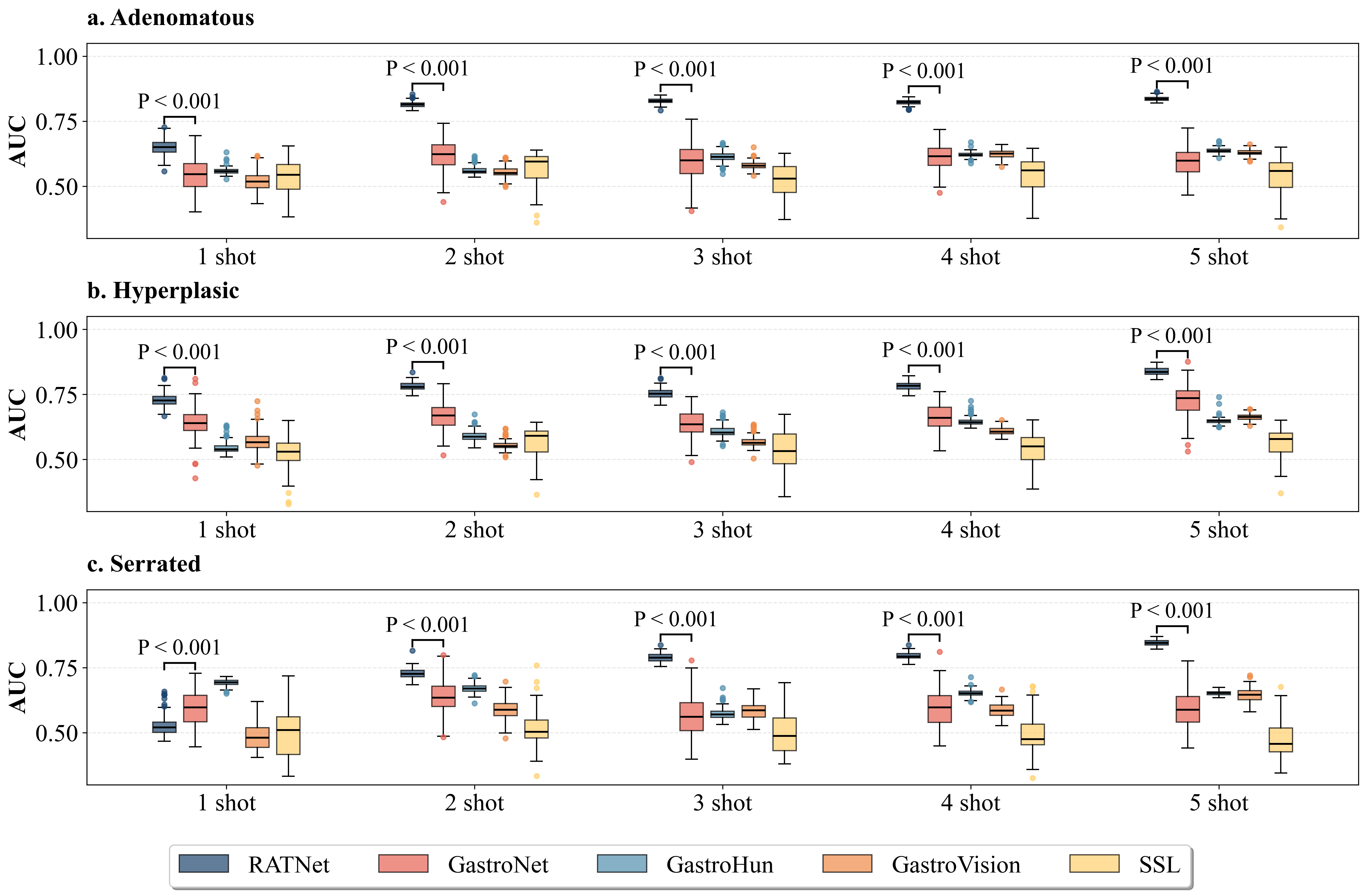}
\caption{Evaluation of the few-shot learning capability for detecting rare conditions. The adenomatous, hyperplastic, and serrated classes from a colonoscopic dataset~\cite{mesejo2016computer} were used to simulate a low-data scenario for rare disease detection. The adaptability and robustness of five pre-trained models were evaluated under a k-shot learning setting. The box plot illustrates the distribution of AUC scores across 100 experimental runs, showing the median (center line), interquartile range (box boundaries at the 25th and 75th percentiles), whiskers (1.5 times the interquartile range), and outliers as individual points. P-values from two-sided independent t-tests are reported in the figure.
}\label{fig:fewshot}
\end{figure}

In this study, we selected three lesion categories that are underrepresented in the colonoscopy dataset: serrated adenomas, hyperplastic lesions, and conventional adenomas. None of these categories appeared in any pretraining corpus, ensuring that the model had no exposure to them during pretraining. For each trial, we randomly sampled k examples from each of the three categories and used them to train a linear classifier. Model performance was then evaluated on a held out colonoscopy test set. Figure \ref{fig:fewshot} shows box and whisker plots summarizing the distribution of AUC scores across 100 repeated experiments for all models.
Across most experiments, RATNet achieved higher median and maximum AUC scores with a smaller interquartile range compared to the baseline model GastroNet~\cite{boers2024foundation}, highlighting its superior performance. As shown in Figure \ref{fig:fewshot}(a) and (c), all models yielded AUC scores below 70\% in the 1-shot scenario. This reflects the inherent difficulty of detecting rare diseases after training on extremely limited data.
These results underscore RATNet's effective capture of discriminative visual patterns, enabling dependable recognition of sparse gastrointestinal abnormalities and highlighting its suitability for deployment in annotation-limited diagnostic settings.

\subsection{Transfer to new sites without training}\label{subsec23}

Distribution shifts arising from differences in patient demographics, imaging devices, or acquisition parameters across institutions frequently compromise the robustness and predictive performance of deep learning systems in medical imaging analysis.
While approaches such as transfer learning and domain adaptation can mitigate these distribution mismatches, they typically necessitate labeled examples from the target environment for model adjustment, constraining the broader applicability and efficient scaling of large-scale pre-trained architectures.

\begin{figure}[htbp]
\centering
\includegraphics[width=0.95\textwidth]{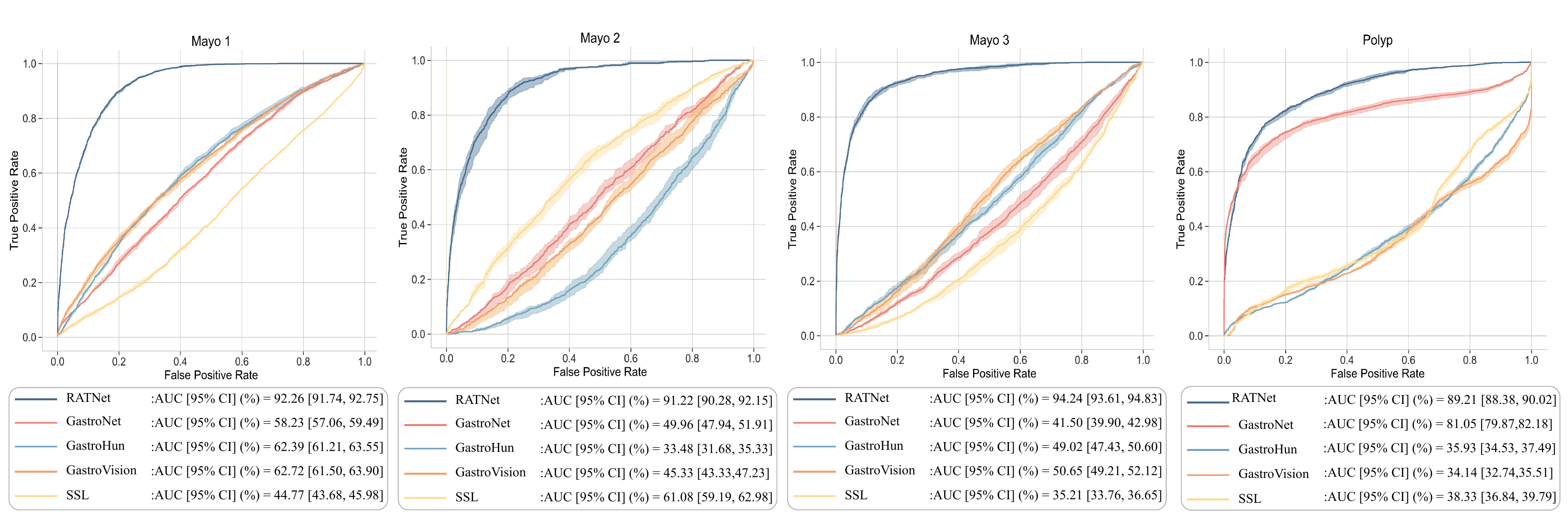}
\caption{Performance on recognizing common gastrointestinal diseases in new  settings without training. The generalizability and robustness of RATNet were evaluated via zero-shot transfer on two unseen datasets (PolypGen~\cite{ali2023multi} and Shaoyifu) and four disease categories previously observed during pre-training (Mayo1, Mayo2, Mayo3, and Polyp). Although these two datasets were not included in pre-training, RATNet had previously encountered the four disease categories in other datasets used for pre-training, specifically Mayo1, Mayo2, and Mayo3 from LIMUC~\cite{polat2023improving}, DAPING, and HyperKvasir~\cite{borgli2020hyperkvasir}, and Polyp from CP-CHILD~\cite{wang2020improved} and Kvasir~\cite{pogorelov2017kvasir}. Therefore, predictions for the four diseases were obtained directly from the pre-trained RATNet head without further fine-tuning, which we refer to as zero-shot transfer. For comparison, two supervised models (GastroHUN~\cite{bravo2025gastrohun} and Gastrovision~\cite{jha2023gastrovision}) and two self-supervised models (GastroNet~\cite{boers2024foundation} and SSL~\cite{bravo2025self}) were evaluated under the same conditions. CI, confidence interval of the ROC curves.}\label{fig:zeroshot}
\end{figure}

To investigate RATNet's capacity for applying acquired pre-training representations to novel clinical settings without additional optimization, we conducted experiments on a new public dataset named PolypGen~\cite{ali2023multi} and a new private dataset named Shaoyifu.
Originating from independent institutions, these datasets encompass varied diagnostic objectives, including polyp identification and Mayo endoscopic subscore assessment.
By leveraging the domain similarity between the current task and all training datasets, RATNet weighted and aggregated the prediction results of all existing multi-task prediction heads based on the target categories. This approach enabled inference in a fully zero-shot configuration, without any task-specific adaptation.
Benchmarking involved four additional pre-trained architectures assessed under equivalent zero-shot protocols.

As illustrated in Figure \ref{fig:zeroshot} through ROC curves inclusive of 95\% confidence bands, the zero-shot outcomes for clinical target identification reveal RATNet's notable AUC achievements: 92.26\% (Mayo 1), 91.22\% (Mayo 2), 94.24\% (Mayo 3), and 89.21\% (polyp identification)—substantially exceeding all comparators.
Such outcomes highlight RATNet's robust extension of learned features to unfamiliar sites and heterogeneous diagnostic applications in the absence of retraining, affirming its suitability for practical clinical integration.

\subsection{Handle long-tailed gastrointestinal diseases}\label{subsec24}

In gastrointestinal endoscopy, abnormality detection commonly faces significant challenges due to imbalanced class prevalence. Prevalent pathologies overwhelmingly dominate clinical observations, whereas uncommon disorders are infrequently encountered, leading to highly uneven data distributions. This disparity creates substantial difficulties in model training, as algorithms often emphasize dominant categories, resulting in reduced sensitivity to infrequent conditions and potential oversight of critical rare findings. Moreover, the paucity of examples for less common diseases heightens the risk of overfitting, thereby limiting effective generalization to new cases. As a result, achieving robust performance and strong adaptability under imbalanced conditions serves as a vital benchmark for assessing advanced foundation models.

\begin{figure}[htbp]
\centering
\includegraphics[width=0.95\textwidth]{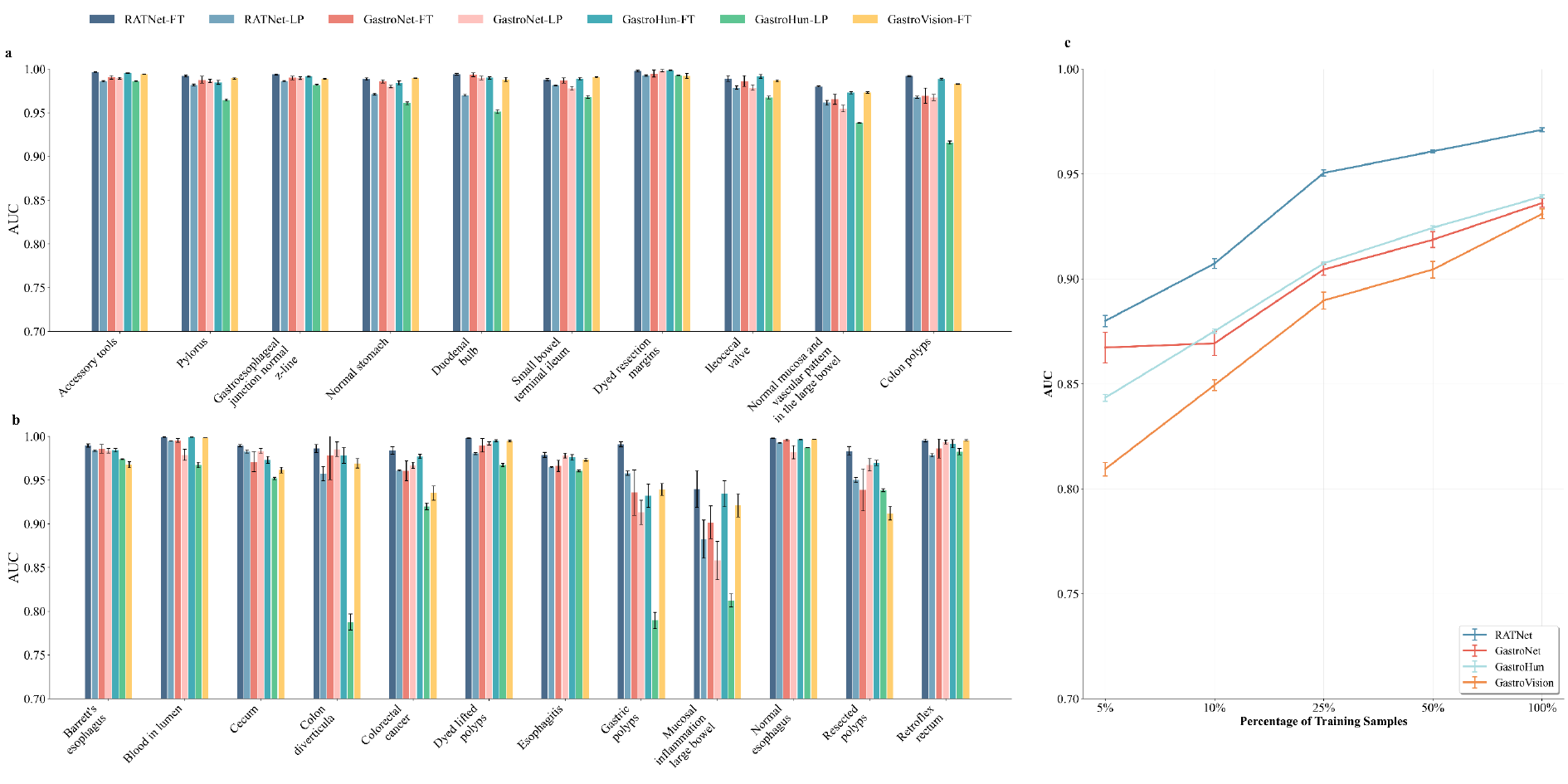}
\caption{Performance for long-tailed gastrointestinal diseases. We evaluated four pre-trained models (GastroHUN~\cite{bravo2025gastrohun}, GastroVision~\cite{jha2023gastrovision}, GastroNet~\cite{boers2024foundation}, and RATNet) on the unseen GastroVision dataset~\cite{jha2023gastrovision} for diagnosing 22 gastrointestinal diseases with a long-tailed distribution. Each model and setup was tested over 10 independent runs. Performance is reported in terms of mean AUC scores with standard deviation error bars for (\textbf{a}) 10 head classes and (\textbf{b}) 12 tail classes. \textbf{c.} Performance of GastroHUN, GastroVision, GastroNet, and RATNet across all 22 diseases for various fractions of training data. To assess label efficiency, we compared the linear probing performance of the models under reduced data settings. For the results reported on each dataset, the appendix 'FT' denotes results obtained via fine-tuning, while the appendix 'LP' denotes results obtained via linear probing.}\label{fig:longtail}
\end{figure}

To investigate RATNet's capabilities in such imbalanced settings, we utilized the independent public GastroVision dataset~\cite{jha2023gastrovision}. This resource includes a broad assortment of categories—ranging from anatomical landmarks and diverse pathologies to post-polypectomy scenarios and routine negative observations—thus capturing the full diversity of real-world endoscopic encounters. RATNet was evaluated on the GastroVision dataset, which contains 22 classes with a long-tailed disease distribution. We assessed its performance under both fine tuning (FT) and linear probing (LP) settings. The results were benchmarked against the fine tuning and linear probing performance reported for GastroNet~\cite{boers2024foundation}, GastroHUN~\cite{bravo2025gastrohun}, and GastroVision~\cite{jha2023gastrovision}.

Figures \ref{fig:longtail}(a) and \ref{fig:longtail}(b) illustrate RATNet's superior accuracy on both high-frequency (head) and low-frequency (tail) conditions relative to the baselines. In particular, linear probing produced an average AUC of 97.11$\pm$0.09\% across 22 conditions for RATNet, clearly outperforming three other foundation models, including GastroNet (93.62$\pm$0.20\%), GastroHUN (93.93$\pm$0.07\%), and GastroVision (93.10$\pm$0.23\%). Extended Data Table 3 provides a more comprehensive evaluation that includes additional aggregated metrics such as mean MCC, average precision, and F1 scores across all classes.
To further probe data efficiency, we performed experiments with progressively reduced training subsets (50\%, 25\%, 10\%, and 5\% of the full data) under the linear probing protocol, comparing RATNet against five alternative pre-trained architectures. As shown in Figure \ref{fig:longtail}(c), RATNet maintained a clear advantage in resource-constrained conditions, especially versus GastroNet. Under identical linear probing setups, RATNet achieved average AUC gains of approximately 1.27\%, 3.8\%, and 4.6\% at the 5\%, 10\% and 25\% data levels, respectively.

\subsection{Respond to novel diseases}\label{subsec25}

Assessing a foundation model’s ability to adapt to novel diseases remains essential for providing reliable and robust diagnostic support. Therefore, we assessed the adaptability of RATNet for identifying gastrointestinal diseases in a video capsule endoscopy setting using the independent public Kvasir-Capsule dataset~\cite{smedsrud2021kvasir} and compared the results with those from five pre-trained models.
To explore RATNet's potential for domain extension, we conducted incrementally and continually pre-training on the Kvasir-Capsule training split, yielding an enhanced variant designated RATNet+KC.

\begin{figure}[htbp]
\centering
\includegraphics[width=0.9\textwidth,height=0.9\textheight]{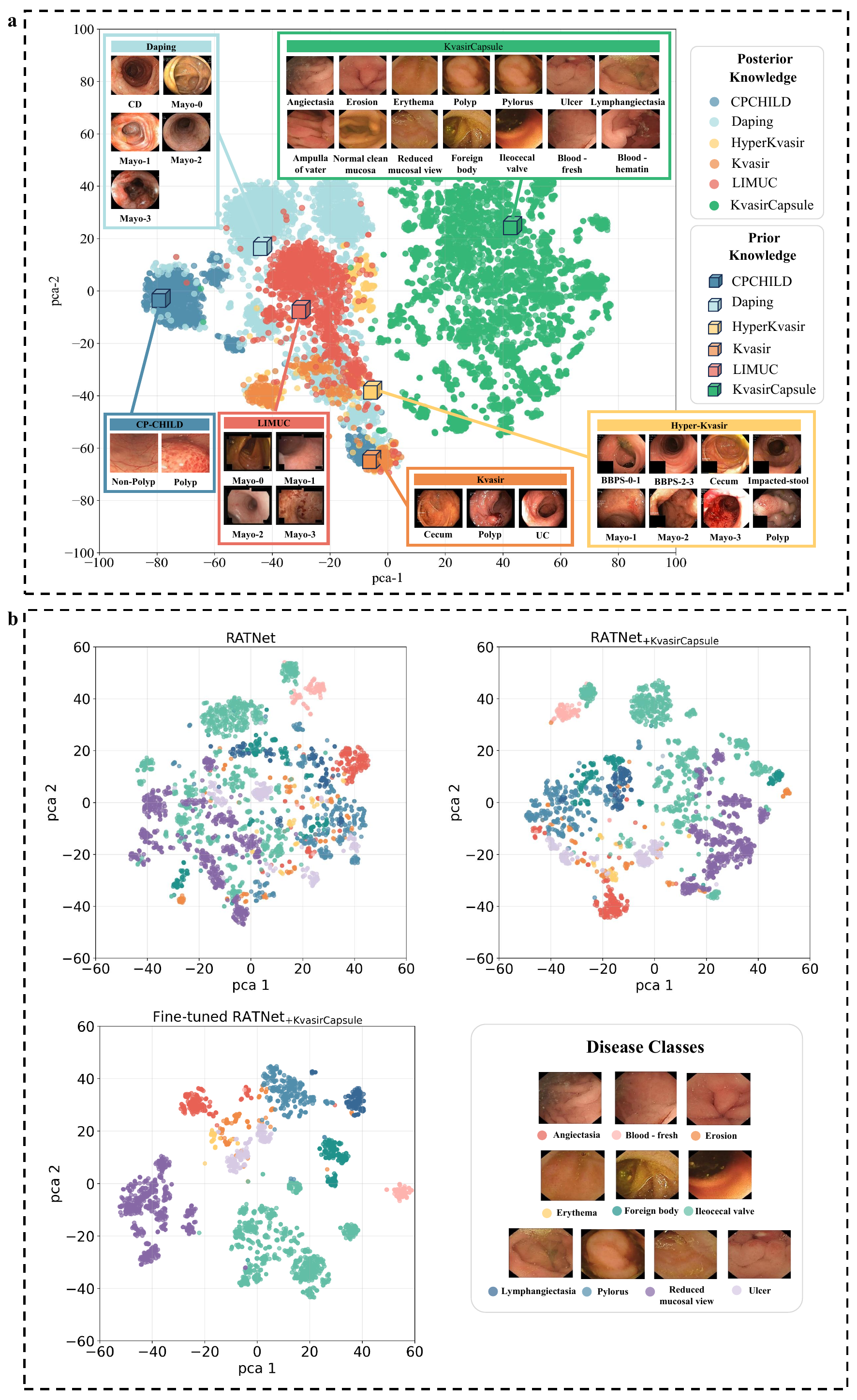}
\caption{Illustration of how the embeddings for Kvasir-Capsule~\cite{smedsrud2021kvasir} dataset's diseases evolve in t-SNE~\cite{van2008visualizing}. \textbf{a.} We progressively enhanced RATNet by continually pre-training it on the Kvasir-Capsule diagnostic task, resulting in an upgraded model termed RATNet\textsubscript{+KC}. The t-SNE visualization illustrates the ability of RATNet\textsubscript{+KC} to capture discriminative prior knowledge for the new task.
\textbf{b.} We demonstrate the evolution of feature separability across different disease categories as the model advances from the pre-trained RATNet to the incrementally learned RATNet\textsubscript{+KC}, and finally to the Fine-tuned RATNet\textsubscript{+KC}.}\label{fig:continue}
\end{figure}

As presented in Extended Data Table 4, the incrementally trained RATNet+KC model attained an AUC of 99.03\%, exceeding the performance of the standard RATNet model and consistently outperforming the other four pre-trained models. For deeper insight into these improvements, we generated t-distributed stochastic neighbor embedding (t-SNE)~\cite{van2008visualizing} plots of feature representations from the Kvasir-Capsule test split using both RATNet and RATNet+KC (Figure \ref{fig:continue}). Post-refinement embeddings revealed clearer clustering of capsule-specific pathologies apart from features learned on other modalities, indicating stronger capture of domain-relevant patterns for precise classification.
Together, these findings highlight RATNet's robust generalization and data-efficient extension to out-of-domain applications, including rare or emerging conditions that may appear in future clinical scenarios. With effective continued pre-training and strong baseline performance, RATNet emerges as a flexible resource for evolving diagnostic demands.

\subsection{Protect privacy and distribute pretraining}\label{subsec26}

RATNet typically undergoes centralized pretraining on public datasets, without addressing the privacy concerns that arise in cross-center collaborations. In contrast, constructing expansive multimodal AI systems for healthcare relies on large volumes of sensitive clinical records, requiring strict safeguards for patient confidentiality and distributed training across institutions.
We introduce Federated RATNet to enable privacy-preserving collaboration, instantiating client models at each participating center (see Figure \ref{fig:overview}(c)). These clients optimize independently using their local data, after which a coordinator gathers the updated parameters, performs parameter averaging, and redistributes the global model for subsequent training cycles.
We simulate a federated learning scenario by distributing different pre-training datasets across three local sites, each hosting an instance of RATNet. In the simulation, while some sites (such as Site 2 and Site 3 in Figure \ref{fig:overview}(c)) utilize public datasets, all local data at each site are treated as private and remain inaccessible to other sites. This approach ensures that each site maintains the confidentiality of its data throughout the training process.
Results in Extended Data Table 5 highlight Federated RATNet's robustness to varying annotation quality among clients, which is a challenge that is often overlooked in traditional distributed methods. The results also demonstrate significant performance improvements from cross-node knowledge integration compared to standalone local optimization.
Ultimately, this framework supports secure international partnerships on proprietary medical data, accelerating innovation in open medical foundation models while upholding patient privacy standards.

\section{Discussion}\label{sec3}

This paper presents RATNet, a foundation model for gastrointestinal endoscopic imaging, which acquires and transfers relevant knowledge embedded in heterogeneous expert annotations from five gastrointestinal endoscopy datasets via cyclic pre-training. Evaluated across six clinical scenarios, RATNet demonstrates superior generalizability, adaptability, robustness, and scalability, outperforming four existing foundation models in diagnosing gastrointestinal diseases.

RATNet is generalizable. It outperforms other pre-training models across five internal test datasets, highlighting its broad applicability in diagnosing common gastrointestinal diseases (Figure \ref{fig:diagnosis}(b) and Extended Data Table 2). Analogical reasoning is a key manifestation of clinical reasoning in diagnostic decision-making: physicians draw on medical knowledge and prior clinical experience to establish similarities between current patient signs and previous cases, predict unobserved signs that may emerge, and infer connections among these signs to reach a diagnosis~\cite{jia2020patient}. However, when dealing with numerous or complex features (such as those in medical images), this approach becomes particularly challenging as it is difficult to quantify such comparisons~\cite{gruger2024enhancing}. RATNet addresses this by employing abstract representations that preserve perceptual richness (via image embedding into vectors) to acquire and transfer relevant knowledge, enabling the model to adapt to analogical reasoning under high-dimensional and uncertain inputs. This reasoning mode enhances the model's generalization, allowing it to perform robustly across different domains for the same disease (Figure \ref{fig:diagnosis}(c))

RATNet is adaptable. While humans can solve visual reasoning puzzles requiring logic with only a few or even zero examples, existing deep learning models like GastroNet~\cite{boers2024foundation} lack this capacity for analogical reasoning and still require extensive training data to achieve similar performance on the same task~\cite{kim2020few}. In this work, we address this few-shot (or zero-shot) visual reasoning problem via an analogical reasoning framework. Specifically, RATNet abstracts knowledge from multiple base tasks to build a repository of isolated experiences, each summarizing insights from distinct tasks (Figure \ref{fig:diagnosis}(a)). During analogical reasoning, the model retrieves relevant knowledge by assessing similarities in both content and structure between the target domain and familiar base domains within this repository~\cite{doumas2022theory}. Based on the degree of similarity, knowledge from the base domains is then transferred to enrich the understanding of the target domain, ultimately guiding the final diagnosis. RATNet outperforms GastroNet in detecting rare diseases using only 1\textasciitilde5 training samples (Figure \ref{fig:fewshot}), underscoring its clinical utility in data-scarce environments. Furthermore, RATNet demonstrates remarkable adaptability to variations in diagnostic settings, achieving high AUC scores in a zero-shot transfer learning scenario for detecting four diseases absent from the training data (Figure \ref{fig:zeroshot}). This ability to generalize to new sites or tasks without requiring retraining highlights its significant potential for real-world deployment.

RATNet is robust. Gastrointestinal diseases often exhibit multi-lesion characteristics~\cite{borgli2020hyperkvasir}, leading to a long-tailed distribution of disease incidence. This property frequently results in highly imbalanced class representations in endoscopic imaging datasets. Due to the instance imbalance between majority and minority classes, deep learning methods tend to develop a bias toward majority classes, which limits the generalizability of classifiers and compromises their ability to accurately recognize rare diseases with limited data~\cite{jin2024few}. In addressing the challenge of long-tailed disease distribution, RATNet outperforms other foundation models on the GastroVision dataset~\cite{jha2023gastrovision}, which exhibits typical long-tail characteristics (Extended Data Table 3). Under limited data scenarios, RATNet consistently demonstrates more robust performance than GastroNet (Figure \ref{fig:longtail}(c)), underscoring the importance of knowledge acquisition and transfer. The analogical reasoning mechanism, inspired by human cognition, enables the model to generate discriminative features that mitigate overfitting to head classes and effectively suppress model bias, thereby enhancing recognition of tail classes.

RATNet is extensible. When incorporating new capsule endoscopy diagnostic tasks, RATNet successfully acquires knowledge from these tasks through its incremental learning capability (Figure \ref{fig:continue}(a)), effectively separating embeddings representing different diseases (Figure \ref{fig:continue}(b)), thereby revealing its newly acquired ability to capture task-specific features unique to capsule endoscopy. This capability can be further enhanced through fine-tuning to achieve more distinctive embeddings. Notably, RATNet not only responds effectively to new tasks via incremental learning but also exhibits performance improvements on original tasks (Extended Data Table 4), highlighting its potential for extension to emerging diseases and novel diagnostic modalities. The training of foundation models typically requires large-scale, centralized datasets. In healthcare, however, medical data are often siloed across institutions due to privacy concerns, regulatory restrictions, and legal, ethical, and technical barriers to data sharing, making centralized data aggregation particularly challenging~\cite{ngiam2019big}. To address this issue, we extend RATNet within a federated learning (FL) framework. In this setting, referred to as Federated RATNet, local models are pre-trained using the isolated data from each medical center. A shared model is then aggregated centrally and redistributed to local sites for fine-tuning. The resulting Federated RATNet model achieves performance comparable to the version trained on a centralized dataset and outperforms RATNet models trained solely on single-institution data (Extended Data Table 4), aligning with findings from recent studies~\cite{sheller2018multi,sheller2020federated}. Furthermore, the multi-task head and knowledge repository design of RATNet overcome the limitations of conventional FL when dealing with heterogeneity~\cite{wahab2024federated} in labels and data sources across different centers. By facilitating collaborative analysis across multiple data silos through the exchange of model updates rather than raw data, RATNet can harness the potential of globally distributed healthcare data from diverse populations, uncovering insights inaccessible to isolated institutions~\cite{li2025challenges}. This makes RATNet a privacy-preserving foundation model without compromising its strong generalization capabilities.

RATNet is open, cognitive and affordable. Its openness enables researchers to perform fine-tuning and incremental learning on new datasets, thereby enhancing diagnostic accuracy across diverse clinical scenarios. All training data and annotations are derived from diverse datasets. Through its analogical reasoning mechanism, RATNet efficiently acquires and transfers knowledge across different domains, eliminating the need for manual unification of heterogeneous annotations. This significantly reduces data acquisition and processing costs while enhancing the model’s cross-domain cognitive capabilities. Looking forward, we plan to extend RATNet's capabilities beyond classification to encompass localization~\cite{senthil2024benchmarking}, segmentation~\cite{saravanan2024benchmarking}, and their integration~\cite{islam2025foundation}. Furthermore, we anticipate that its public accessibility will attract more diverse data sources, particularly from underrepresented regions and populations with limited healthcare resources. This will enable the assessment of model bias across demographic attributes including gender, race, age, and other population factors, thereby promoting equitable and accurate diagnosis across populations, addressing ethical concerns in clinical AI, and ensuring more comprehensive and inclusive robustness evaluation.

In conclusion, RATNet demonstrated impressive performance in diagnosing gastrointestinal diseases across various scenarios, including common conditions, rare diseases with few samples, long-tailed distributions, zero-shot transfer to new sites, response to novel diseases, and privacy-preserving federated learning. Its generalizability, adaptability, robustness, scalability, openness, and affordability make it a powerful foundation model for gastrointestinal endoscopic imaging, capable of democratizing expert-level diagnostics. The model's exceptional capabilities are attributable to its innovative analogical reasoning framework, which cyclically acquires and transfers knowledge from heterogeneous expert annotations across multiple gastrointestinal endoscopy datasets, effectively addressing challenges like data scarcity, domain shift, and label heterogeneity without requiring costly label unification. RATNet represents a significant shift from task-specific learning toward generalized, multi-center medical intelligence. By promoting collaborative efforts and open-source accessibility, it not only enhances clinical efficiency but also facilitates equitable access to AI-assisted diagnostics in low-resource settings. Future research directions include extending the analogy reasoning mechanism to localization and segmentation, incorporating temporal reasoning into endoscopic video stream analysis, and exploring multimodal alignment among histopathology, imaging, and genomic data. These steps will accelerate the development of a truly universal gastrointestinal intelligence model, capable of supporting precision gastroenterology and advancing global medical democratization.

\section{Methods}\label{sec4}

\subsection{Model architecture}\label{subsec41}

\begin{figure}[t]
\centering
\includegraphics[width=0.9\textwidth]{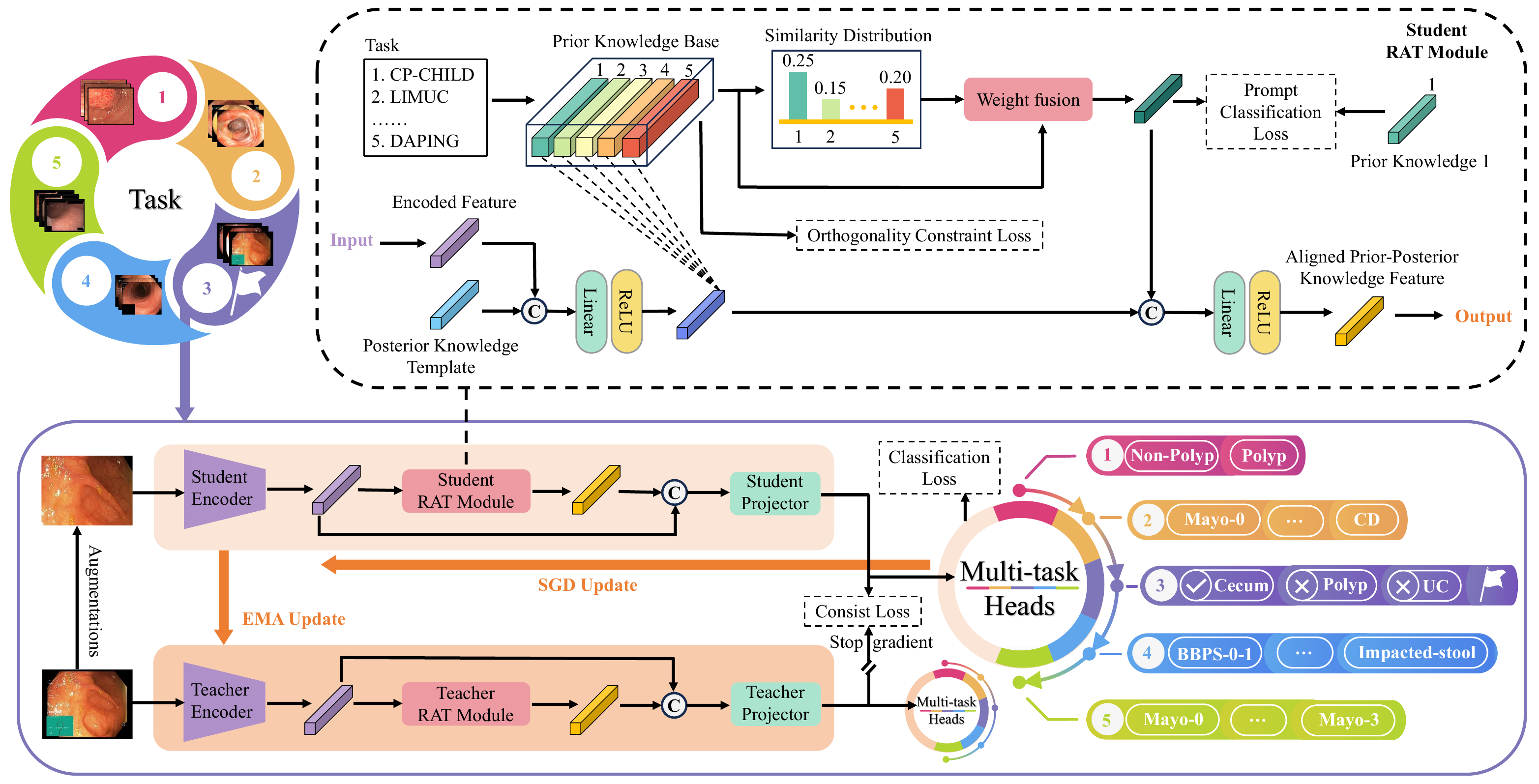}
\caption{Illustration of the training pipeline of RATNet. RATNet is a teacher-student framework with multi-task heads designed to address heterogeneous label spaces across tasks. Specifically, each task is associated with an independent prediction head that maps shared representations to its own label set, while cyclical pretraining enables knowledge acquisition and transfer across tasks despite label inconsistency. In each iteration, the student scans the dataset task by task, and the relevance-knowledge acquisition and transfer module updates both the prior knowledge base and the posterior knowledge template. The student’s newly acquired knowledge is accumulated into the teacher through exponential moving average (EMA), allowing the teacher to guide the student on subsequent tasks. After pretraining, the knowledge accumulated in the teacher can be transferred to downstream target tasks.}\label{fig:RATNet}
\end{figure}

We propose a framework named relevance-knowledge acquisition and transfer network (RATNet), which is designed to acquire task-related knowledge from heterogeneous expert annotations across multiple tasks, thereby constructing a prior knowledge base. When encountering a new task not covered by the original training data, RATNet enables the transfer of relevant knowledge from this base to the new task. Specifically, it retrieves associated factual information related to the new task from the prior knowledge base and grounds the network’s responses on such information, thereby producing more accurate and relevant outputs. As illustrated in Fig.~\ref{fig:RATNet}, RATNet consists of an encoder, a relevance-Knowledge acquisition and transfer (RAT) module, a projector, and multiple task heads. It is built upon a teacher–student architecture capable of handling heterogeneous labels across tasks and is trained using a cyclic pre-training strategy.

\subsubsection{Teacher-student architecture for heterogeneous label}\label{subsubsec411}

To cope with heterogeneous labels across multi-dataset pretraining, we avoid forcibly unifying all annotations into a single shared label set. Instead, we treat heterogeneity as a source of multi-task expertise: each dataset/task is equipped with a plug-in task head, allowing the model to learn the native semantics of each annotation scheme on top of a shared visual encoder. To reduce gradient interference that arises when summing losses from incompatible label spaces, we adopt a cyclic pretraining schedule where the student updates on one task at a time and revisits tasks repeatedly to limit forgetting. In parallel, a teacher model with the same architecture accumulates the student’s task-specific knowledge via EMA after each task/epoch, gradually integrating cross-task experience. A student–teacher consistency loss, implemented with projection heads that map representations into a common feature space, feeds the teacher’s stable, cross-heterogeneity knowledge back to the student as additional supervision, helping preserve transferable representations while learning new label spaces.

\subsubsection{Relevance-knowledge acquisition and transfer module}\label{subsubsec412}

The objective of this module is to enable the model to acquire prior knowledge from each task, with all such knowledge collectively forming a knowledge base. When presented with a new task, the model should be able to retrieve relevance-knowledge from the knowledge base by evaluating the similarity between samples from the new task and each stored knowledge entry. Based on these similarity scores, the model then extracts and integrates the factual information from relevance-knowledge to enable effective cross-task knowledge transfer. Specifically, we begin by randomly initializing a set of learnable knowledge base entries $KB \in \mathbb{R}^{T \times E} = \left\{{b}_1, {b}_2, \ldots, {b}_T \right\}$, where $T$ corresponds to the total number of tasks in the pre-training set. To facilitate the learning of distinct task-specific knowledge, we impose an orthogonality constraint on $KB$ to facilitate that each knowledge vector retains its unique task-related characteristics.

Since the pre-training dataset may not encompass all possible tasks, and the model may encounter unseen tasks during inference, relying solely on $KB$ is insufficient. We aim to extract knowledge from $KB$ that is relevant to the current task. Therefore, we develop a posterior knowledge generation mechanism that projects the current encoded features into a posterior knowledge vector space, enabling adaptive generation of posterior knowledge rather than relying exclusively on the prior domain distribution in $KB$. To achieve this, we randomly initialize a learnable posterior knowledge template, denoted as $t_{pk}$. Given the encoded feature $v_e$, we concatenate $v_e$ with ${t_pk}$ and feed $[v_e \copyright t_{pk}]$ into an MLP for fusion, thereby generating dynamic posterior knowledge features ${k_p}$.

To fully leverage the prior knowledge from $KB$ and the posterior knowledge captured by ${k_p}$, we propose a knowledge alignment and fusion strategy to seamlessly integrate these two types of knowledge. Considering that the domain knowledge corresponding to different tasks in $KB$ may have varying degrees of relevance to the current task, we aim to selectively combine this information based on relevance to generate guidance information aligned with the posterior knowledge for new tasks in unseen domains. Specifically, for each prior knowledge entry $b_i \in KB$, we compute its cosine similarity with the posterior knowledge ${k_p}$ and apply a softmax function to obtain adaptive weights $\mathcal{W} = \left\{\omega_1, \omega_2, \ldots, \omega_Z \right\}$:
\begin{equation}
	sim_i = \frac{{k_p \cdot b_i}}{{\left\| {k_p} \right\| \cdot \left\| {b_i} \right\|}} .\label{eq1}
\end{equation}
\begin{equation}
	\omega_i = \frac{{exp(sim_i) / \tau}}{\sum\limits_{k=1}^{T} {exp(sim_k) / \tau}} .\label{eq2}
\end{equation}

Subsequently, the weighted aggregation of the prior knowledge in $KB$ is computed as follows:
\begin{equation}
k_a = \sum\limits_{i=1}^{T} \omega_i \cdot {b}_i.\label{eq3}
\end{equation}

During pre-training, the resulting aligned prior knowledge feature $k_a$ is constrained using a task similarity loss $\mathcal{L}_{ts}$. This loss is defined by measuring the similarity between the prior knowledge $b_i$ of the current task in $KB$ and the aligned prior knowledge $k_a$:
\begin{equation}
\mathcal{L}_{ts} = (1-\frac{{k_a \cdot b_i}}{{\left\| {k_a} \right\| \cdot \left\| {b_i} \right\|}})^2.\label{eq4}
\end{equation}

Finally, we concatenate the posterior knowledge ${k_p}$ with ${k_a}$ to form the combined feature $[{k_p} \copyright {k_a}]$. This concatenated feature is then passed through an MLP block for fusion, producing the final aligned prior-posterior knowledge feature.

\subsection{Implementation details}\label{subsec42}

RATNet employs a Swin large backbone with an input resolution of 384×384. Both the teacher and student encoders, along with the projector and multi-task head, are initialized with weights pre-trained on our dataset. The model was trained using a stochastic gradient descent optimizer with an initial learning rate of 0.003 and a batch size of 64 on an NVIDIA H100 GPU with 80 GB of memory. At the end of each task, the teacher model was updated using an exponential moving average of the student’s weights over one epoch with a momentum of 0.9. Image augmentations included random cropping and rotation, as well as variations in brightness, contrast, and gamma distribution. The model underwent 100 epochs of pre-training, completing 100 iterations over the entire dataset.

\subsection{Datasets for developing and evaluating proposed model}\label{subsec43}

\subsubsection{Pretraining datasets}\label{subsubsec431}

Our proposed model was pre-trained on a total of 38,995 gastrointestinal endoscopy images obtained from five datasets provided by institutions worldwide, all of which were annotated by domain experts. Each dataset underwent uniform preprocessing and resampling procedures to form a consolidated pre-training dataset.
The CP-CHILD dataset~\cite{wang2020improved} contains colonoscopy images collected from 1,600 children in Hunan Province, including 7,874 normal images and 1,395 polyp images. All images were obtained at the Hunan Children’s Hospital between March 2018 and April 2019.
The LIMUC dataset~\cite{polat2023improving} includes 11,276 images and 1,043 colonoscopy videos from 564 patients, acquired at the Department of Gastroenterology, Marmara University School of Medicine, between December 2011 and July 2019. After resampling, 10,765 images were retained. These images are categorized into four classes according to the Mayo endoscopic scoring system: Mayo 0 to 3.
The Kvasir dataset~\cite{pogorelov2017kvasir} consists of images acquired from the gastrointestinal tract using endoscopic equipment at Vestre Viken Health Trust, Norway. We sampled 986 cecum images, 1,000 polyp images, and 1,000 ulcerative colitis images.
The HyperKvasir dataset~\cite{borgli2020hyperkvasir} was collected from gastroscopies and colonoscopies performed at Bærum Hospital, Norway between 2008 and 2016. We selected 4,713 gastrointestinal images across eight categories: cecum, BBPS-0-1, BBPS-2-3, impacted stool, polyp, Mayo-1, Mayo-2, and Mayo-3.
The Daping dataset was prospectively collected during routine clinical examinations at Daping Hospital in China and includes 21,011 images labeled into five categories: Mayo-0, Mayo-1, Mayo-2, Mayo-3, and CD (Crohn’s disease).
From the combined data, 34,120 images were allocated for pre-training and 4,875 images were reserved for validation.

\subsubsection{Data for few-shot learning}\label{subsubsec432}

The Colonoscopic dataset~\cite{mesejo2016computer} consists of 76 colonoscopy videos and covers three lesion types: serrated adenomas, hyperplastic lesions, and adenomas.
These three lesion types were used in a few-shot linear-probing benchmark to simulate the challenging scenario of rare diseases, in which a hospital has access to only a limited number of cases for model training. 

\subsubsection{Data for zero-shot transfer}\label{subsubsec433}

PolypGen~\cite{ali2023multi} is a general-purpose dataset for polyp segmentation and detection. It consists of 8,037 frames, including both single images and sequences. The comprehensive dataset includes 3,762 positive frames and 4,275 negative frames collected from six different centers across Europe and Africa. From this collection, we sampled 3,240 positive frames and 2,194 negative frames for our experiments. The Shaoyifu Dataset was collected by the Department of Gastroenterology at Shaoyifu Hospital, Zhejiang University School of Medicine. It comprises 9,620 images categorized into four classes: Mayo-1, Mayo-2, Mayo-3, and CD (Crohn’s Disease).

\subsubsection{Data in long-tailed distributions}\label{subsubsec434}

The GastroVision dataset~\cite{jha2023gastrovision} was collected from two medical centers: the Department of Gastroenterology at Bærum Hospital, Vestre Viken Hospital Trust in Norway, and Karolinska University Hospital in Sweden, using standard endoscopy systems provided by Olympus Europe and Pentax Medical Europe, both based in Germany. It comprises 8,000 images annotated with 27 distinct categories.
These categories are further divided into two major groups: the upper gastrointestinal tract and the lower gastrointestinal tract. The sample distribution across categories is imbalanced, exhibiting a long-tail distribution. The sample distribution across categories is highly imbalanced, exhibiting a pronounced long-tail pattern. Specifically, 10 head categories contain relatively large numbers of samples, ranging from 200 to 1,467 instances per category. In contrast, the remaining 17 tail categories are significantly underrepresented. Each of these tail categories contains only 6 to 171 samples.

\subsubsection{Data for incremental learning}\label{subsubsec435}

The Kvasir-Capsule~\cite{smedsrud2021kvasir} is a large video capsule endoscopy (VCE) dataset collected from examinations at a Norwegian hospital, comprising 43 annotated videos. It includes 47,238 frames annotated with 14 different categories. Unlike similar datasets containing colonoscopy or esophagogastroscopy images, this dataset focuses on the small intestine, which exhibits a distinct mucosal surface characterized by intestinal villi. Furthermore, VCE images have significantly lower resolution and frame rates, the intestine is not insufflated as in conventional endoscopy, different optical systems are used, and the capsule's movement is uncontrolled compared to flexible endoscopes employed during manual procedures.

\subsection{Comparative pretrained models}\label{subsec44}

To evaluate the performance of our RATNet model, we conducted a comprehensive comparison with several existing gastrointestinal diagnostic models, namely: GastroNet~\cite{boers2024foundation}, SSL~\cite{bravo2025self}, GastroHUN~\cite{bravo2025gastrohun}, and GastroVision~\cite{jha2023gastrovision}. The model configurations and training details for all these methods are provided in Extended Data Table 6. 
GastroNet was trained using self-supervised learning on a dataset comprising 5014174 unlabeled gastrointestinal endoscopy images from eight different medical centers. The model was configured with a Vision Transformer~\cite{DBLP:conf/iclr/DosovitskiyB0WZ21} architecture and pre-trained in-domain using DINO~\cite{DBLP:conf/iclr/0097LL000NS23}.
The SSL approach employed a Masked Autoencoder (MAE)~\cite{he2022masked}, which learns feature representations by reconstructing the original input from partially observed data. It was pre-trained on 99,417 unlabeled samples from the HyperKvasir dataset~\cite{borgli2020hyperkvasir}.
GastroHUN utilizes a ConvNeXt~\cite{liu2022convnet} architecture and enhances model performance through a two-stage pre-training strategy. It was pre-trained on a dataset containing 8,834 labeled images from 387 patients, covering 22 anatomical landmarks in the stomach.
GastroVision was pre-trained in a supervised manner using a DenseNet\cite{huang2017densely} architecture on a multi-center, open-access gastrointestinal endoscopy dataset consisting of 8,000 images across 27 classes. MiniGPT-v2~\cite{chen2023minigpt} employs a large language model as a unified interface for handling multiple vision-language tasks and utilizes task identifiers to guide the model in distinguishing between different types of instructions, thereby improving multi-task learning efficiency. LLaVA-v1~\cite{liu2023visual} introduces the concept of "visual instruction tuning," which enhances the model's ability to comprehend and generate visual content through an instruction-based learning mechanism. Building upon LLaVA-v1, LLaVA-v1.5~\cite{liu2024improved} incorporates higher-quality training data and optimized fine-tuning strategies, leading to significantly improved multimodal alignment performance. Bunny-v1.0-3B~\cite{he2024efficient} is a lightweight multimodal large model that achieves strong performance with a limited number of parameters by adopting a flexible vision-language modular architecture and efficient data sampling strategies. MGM-2B~\cite{li2024mini} proposes a general framework designed to enhance multimodal model capabilities, focusing on three key aspects: high-resolution visual token modeling, construction of high-quality training data, and self-guided model generation. MobileVLM-1.7B~\cite{chu2023mobilevlm} features a lightweight and efficiently trained architecture, enabling comprehensive visual-language understanding and response capabilities on mobile devices while maintaining a balance between semantic consistency and inference speed. LLaVA-Med-v1.0 and LLaVA-Med-v1.5~\cite{li2023llava} are large vision-language models specifically designed for medical applications. They leverage biomedical image-text data from PubMed Central and utilize GPT-4 to automatically generate medical question-answer pairs, thereby facilitating effective instruction tuning. ColonGPT is an intelligent model tailored for colonoscopy procedures, integrating linguistic and visual information to assist in the analysis and diagnosis of colonoscopy images. All the aforementioned vision-language models were fine-tuned in accordance with the data split protocol of the ColonINST dataset~\cite{ji2026frontiers}.

\subsection{Evaluation metrics and statistical analysis}\label{subsec45}

To comprehensively assess the model's classification performance, we employed the following metrics, each emphasizing distinct aspects.

The Area Under the Receiver Operating Characteristic Curve (AUC) is a prominent metric in binary classification, quantifying the area beneath the ROC curve that plots True Positive Rate (TPR) against False Positive Rate (FPR) across thresholds. AUC ranges from 0 to 1, with 0.5 denoting random guessing and 1 indicating flawless classification. Higher values reflect superior discriminative capability between classes. AUC excels in providing a threshold-independent overview, facilitating model comparisons on balanced datasets.

The F1 Score harmonizes precision and recall through their harmonic mean, serving as a key metric for binary and multi-class classification. Ranging from 0 to 1, a value of 1 signifies perfect precision and recall, while 0 implies no true positives identified. By addressing both false positives and negatives, F1 Score outperforms accuracy, particularly in imbalanced datasets where it ensures balanced assessment.

Average Precision (AP) is essential for evaluating models in information retrieval and object detection, defined as the area under the Precision-Recall curve. It measures sustained high precision across recall levels and varying thresholds. In object detection, mean AP (mAP) aggregates category-specific values. AP is ideal for imbalanced scenarios or rare classes, emphasizing positive class prediction accuracy.

The Matthews Correlation Coefficient (MCC), proposed by Brian W. Matthews in 1975, evaluates binary classifiers by incorporating all confusion matrix components: true positives, true negatives, false positives, and false negatives. MCC spans -1 to +1, with +1 for perfect agreement, 0 for random performance, and -1 for complete discord. As a balanced metric, it offers reliable insights into classifier efficacy, even on imbalanced data, surpassing F1 Score or accuracy.

To evaluate the statistical significance of observed differences in model performance and to draw reliable conclusions, we employed a comprehensive statistical analysis approach. This method integrated an independent two-sample t-test with the calculation of 95\% confidence intervals.

In experiments involving fine-tuning and linear-probing, the random initialization inherent to the linear classifier meant that each execution could yield distinct outcomes. Consequently, to ensure the robustness of our findings, we performed at least ten evaluations for each model and reported both the mean and standard deviation for every performance metric. A two-sided independent t-test was subsequently conducted for the statistical analysis.

\backmatter

\bmhead{Supplementary information}

Extended data is available for this paper in Supplementary information.

\bmhead{Acknowledgements}

The work was partially supported by the National Natural Science Foundation of China (62471293), by Chongqing Excellence Program for Innovation and Entrepreneurship Leadership Talent Project (CQYC20220303576), and by the Natural Science Foundation of Chongqing, China (CSTB2024NSCQ-LZX0141).

\bibliography{sn-bibliography}

@inproceedings{xu2022patch,
  title={Patch-level instance-group discrimination with pretext-invariant learning for colitis scoring},
  author={Xu, Ziang and Ali, Sharib and Gupta, Soumya and Leedham, Simon and East, James E and Rittscher, Jens},
  booktitle={International Workshop on Machine Learning in Medical Imaging},
  pages={101--110},
  year={2022},
  organization={Springer}
}

@article{xu2022mayo,
  title={The Mayo endoscopic score is a novel predictive indicator for malignant transformation in ulcerative colitis: a long-term follow-up multicenter study},
  author={Xu, Weimin and Liu, Fangyuan and Tang, Wenbo and Gu, Yubei and Zhong, Jie and Cui, Long and Du, Peng},
  journal={Frontiers in surgery},
  volume={9},
  pages={832219},
  year={2022},
  publisher={Frontiers Media SA}
}

@article{van2008visualizing,
  title={Visualizing data using t-SNE.},
  author={Van der Maaten, Laurens and Hinton, Geoffrey},
  journal={Journal of machine learning research},
  volume={9},
  number={11},
  year={2008}
}

@article{ji2026frontiers,
  title={Frontiers in intelligent colonoscopy},
  author={Ji, Ge-Peng and Liu, Jingyi and Xu, Peng and Barnes, Nick and Khan, Fahad Shahbaz and Khan, Salman and Fan, Deng-Ping},
  journal={Machine Intelligence Research},
  volume={23},
  number={1},
  pages={70--114},
  year={2026},
  publisher={Springer}
}

@article{wang2023global,
  title={Global burden of digestive diseases: a systematic analysis of the global burden of diseases study, 1990 to 2019},
  author={Wang, Yichen and Huang, Yuting and Chase, Robert C and Li, Tian and Ramai, Daryl and Li, Si and Huang, Xiaoquan and Antwi, Samuel O and Keaveny, Andrew P and Pang, Maoyin},
  journal={Gastroenterology},
  volume={165},
  number={3},
  pages={773--783},
  year={2023},
  publisher={Elsevier}
}

@article{arnold2020global,
  title={Global burden of 5 major types of gastrointestinal cancer},
  author={Arnold, Melina and Abnet, Christian C and Neale, Rachel E and Vignat, Jerome and Giovannucci, Edward L and McGlynn, Katherine A and Bray, Freddie},
  journal={Gastroenterology},
  volume={159},
  number={1},
  pages={335--349},
  year={2020},
  publisher={Elsevier}
}

@article{sung2021global,
  title={Global cancer statistics 2020: GLOBOCAN estimates of incidence and mortality worldwide for 36 cancers in 185 countries},
  author={Sung, Hyuna and Ferlay, Jacques and Siegel, Rebecca L and Laversanne, Mathieu and Soerjomataram, Isabelle and Jemal, Ahmedin and Bray, Freddie},
  journal={CA: a cancer journal for clinicians},
  volume={71},
  number={3},
  pages={209--249},
  year={2021},
  publisher={Wiley Online Library}
}

@article{correa2013gastric,
  title={Gastric cancer: overview},
  author={Correa, Pelayo},
  journal={Gastroenterology Clinics of North America},
  volume={42},
  number={2},
  pages={211},
  year={2013}
}

@article{tang2021advances,
  title={Advances in optical gastrointestinal endoscopy: a technical review},
  author={Tang, Yubo and Anandasabapathy, Sharmila and Richards-Kortum, Rebecca},
  journal={Molecular Oncology},
  volume={15},
  number={10},
  pages={2580--2599},
  year={2021},
  publisher={Wiley Online Library}
}

@article{martins2023endoscopic,
  title={Endoscopic imaging for the diagnosis of neoplastic and pre-neoplastic conditions of the stomach},
  author={Martins, Bruno Costa and Moura, Renata Nobre and Kum, Angelo So Taa and Matsubayashi, Carolina Ogawa and Marques, Sergio Barbosa and Safatle-Ribeiro, Adriana Vaz},
  journal={Cancers},
  volume={15},
  number={9},
  pages={2445},
  year={2023},
  publisher={MDPI}
}

@article{tham2025artificial,
  title={Artificial Intelligence in Endoscopy: A Narrative Review},
  author={Tham, CE and Rea, D and Tham, TC},
  journal={The Ulster Medical Journal},
  volume={94},
  number={1},
  pages={16},
  year={2025}
}

@article{xu2025artificial,
  title={Artificial intelligence system improves the quality of digestive endoscopy: A prospective pretest and post-test single-center clinical trial},
  author={Xu, Zewen and Li, Yongrong and Su, Peiqiang and Zhong, Zhuangxia and Zeng, Zuni and Chen, Mingli and Chen, Di and Lan, Cheng},
  journal={Digestive and Liver Disease},
  year={2025},
  publisher={Elsevier}
}

@article{shi2024survey,
  title={A survey on trustworthiness in foundation models for medical image analysis},
  author={Shi, Congzhen and Rezai, Ryan and Yang, Jiaxi and Dou, Qi and Li, Xiaoxiao},
  journal={arXiv preprint arXiv:2407.15851},
  year={2024}
}

@article{zhang2024challenges,
  title={On the challenges and perspectives of foundation models for medical image analysis},
  author={Zhang, Shaoting and Metaxas, Dimitris},
  journal={Medical image analysis},
  volume={91},
  pages={102996},
  year={2024},
  publisher={Elsevier}
}

@article{ma2025fully,
  title={A fully open AI foundation model applied to chest radiography},
  author={Ma, DongAo and Pang, Jiaxuan and Gotway, Michael B and Liang, Jianming},
  journal={Nature},
  pages={1--11},
  year={2025},
  publisher={Nature Publishing Group UK London}
}

@article{wu2025towards,
  title={Towards generalist foundation model for radiology by leveraging web-scale 2d\&3d medical data},
  author={Wu, Chaoyi and Zhang, Xiaoman and Zhang, Ya and Hui, Hui and Wang, Yanfeng and Xie, Weidi},
  journal={Nature Communications},
  volume={16},
  number={1},
  pages={7866},
  year={2025},
  publisher={Nature Publishing Group UK London}
}

@article{wang2024pathology,
  title={A pathology foundation model for cancer diagnosis and prognosis prediction},
  author={Wang, Xiyue and Zhao, Junhan and Marostica, Eliana and Yuan, Wei and Jin, Jietian and Zhang, Jiayu and Li, Ruijiang and Tang, Hongping and Wang, Kanran and Li, Yu and others},
  journal={Nature},
  volume={634},
  number={8035},
  pages={970--978},
  year={2024},
  publisher={Nature Publishing Group UK London}
}

@inproceedings{he2025foundational,
  title={Foundational Multi-Task Multimodal Model for Upper GI Endoscopy},
  author={He, Yuxuan and Chen, Qilei and Liu, Benyuan and Cao, Yu},
  booktitle={Proceedings of the IEEE/CVF International Conference on Computer Vision},
  pages={6612--6621},
  year={2025}
}

@article{zhang2024learning,
  title={Learning to adapt foundation model dinov2 for capsule endoscopy diagnosis},
  author={Zhang, Bowen and Chen, Ying and Bai, Long and Zhao, Yan and Sun, Yuxiang and Yuan, Yixuan and Zhang, Jianhua and Ren, Hongliang},
  journal={Procedia Computer Science},
  volume={250},
  pages={188--194},
  year={2024},
  publisher={Elsevier}
}

@article{dermyer2025endodino,
  title={Endodino: A foundation model for gi endoscopy},
  author={Dermyer, Patrick and Kalra, Angad and Schwartz, Matt},
  journal={arXiv preprint arXiv:2501.05488},
  year={2025}
}

@article{devkota2025federated,
  title={Federated Foundation Model for GI Endoscopy Images},
  author={Devkota, Alina and Amireskandari, Annahita and Palko, Joel and Thakkar, Shyam and Adjeroh, Donald and Jiang, Xiajun and Bhattarai, Binod and Gyawali, Prashnna K},
  journal={arXiv preprint arXiv:2505.24108},
  year={2025}
}

@inproceedings{kondrateva2021domain,
  title={Domain shift in computer vision models for MRI data analysis: an overview},
  author={Kondrateva, Ekaterina and Pominova, Marina and Popova, Elena and Sharaev, Maxim and Bernstein, Alexander and Burnaev, Evgeny},
  booktitle={Thirteenth International Conference on Machine Vision},
  volume={11605},
  pages={126--133},
  year={2021},
  organization={SPIE}
}

@article{ayana2024multistage,
  title={Multistage transfer learning for medical images},
  author={Ayana, Gelan and Dese, Kokeb and Abagaro, Ahmed Mohammed and Jeong, Kwangcheol Casey and Yoon, Soon-Do and Choe, Se-woon},
  journal={Artificial Intelligence Review},
  volume={57},
  number={9},
  pages={232},
  year={2024},
  publisher={Springer}
}

@book{gentner2017analogical,
  title={International handbook of thinking and reasoning},
  author={Ball, Linden J and Thompson, Valerie A},
  year={2017},
  publisher={Routledge}
}

@article{gust2008analogical,
  title={Analogical reasoning: a core of cognition.},
  author={Gust, Helmar and Krumnack, Ulf and K{\"u}hnberger, Kai-Uwe and Schwering, Angela},
  journal={K{\"u}nstliche Intell.},
  volume={22},
  number={1},
  pages={8--12},
  year={2008}
}

@article{gentner1997reasoning,
  title={Reasoning and learning by analogy: Introduction.},
  author={Gentner, Dedre and Holyoak, Keith J},
  journal={American psychologist},
  volume={52},
  number={1},
  pages={32},
  year={1997},
  publisher={American Psychological Association}
}

@book{guallart2014analogical,
  title={Systematic approaches to argument by analogy},
  author={Ribeiro, Henrique Jales},
  volume={25},
  year={2014},
  publisher={Springer}
}

@article{wang2020improved,
  title={An improved deep learning approach and its applications on colonic polyp images detection},
  author={Wang, Wei and Tian, Jinge and Zhang, Chengwen and Luo, Yanhong and Wang, Xin and Li, Ji},
  journal={BMC Medical Imaging},
  volume={20},
  number={1},
  pages={83},
  year={2020},
  publisher={Springer}
}

@article{polat2023improving,
  title={Improving the computer-aided estimation of ulcerative colitis severity according to mayo endoscopic score by using regression-based deep learning},
  author={Polat, Gorkem and Kani, Haluk Tarik and Ergenc, Ilkay and Ozen Alahdab, Yesim and Temizel, Alptekin and Atug, Ozlen},
  journal={Inflammatory Bowel Diseases},
  volume={29},
  number={9},
  pages={1431--1439},
  year={2023},
  publisher={Oxford University Press US}
}

@inproceedings{pogorelov2017kvasir,
  title={Kvasir: A multi-class image dataset for computer aided gastrointestinal disease detection},
  author={Pogorelov, Konstantin and Randel, Kristin Ranheim and Griwodz, Carsten and Eskeland, Sigrun Losada and de Lange, Thomas and Johansen, Dag and Spampinato, Concetto and Dang-Nguyen, Duc-Tien and Lux, Mathias and Schmidt, Peter Thelin and others},
  booktitle={Proceedings of the 8th ACM on Multimedia Systems Conference},
  pages={164--169},
  year={2017}
}

@article{borgli2020hyperkvasir,
  title={HyperKvasir, a comprehensive multi-class image and video dataset for gastrointestinal endoscopy},
  author={Borgli, Hanna and Thambawita, Vajira and Smedsrud, Pia H and Hicks, Steven and Jha, Debesh and Eskeland, Sigrun L and Randel, Kristin Ranheim and Pogorelov, Konstantin and Lux, Mathias and Nguyen, Duc Tien Dang and others},
  journal={Scientific data},
  volume={7},
  number={1},
  pages={283},
  year={2020},
  publisher={Nature Publishing Group UK London}
}

@article{mesejo2016computer,
  title={Computer-aided classification of gastrointestinal lesions in regular colonoscopy},
  author={Mesejo, Pablo and Pizarro, Daniel and Abergel, Armand and Rouquette, Olivier and Beorchia, Sylvain and Poincloux, Laurent and Bartoli, Adrien},
  journal={IEEE transactions on medical imaging},
  volume={35},
  number={9},
  pages={2051--2063},
  year={2016},
  publisher={IEEE}
}

@article{ali2023multi,
  title={A multi-centre polyp detection and segmentation dataset for generalisability assessment},
  author={Ali, Sharib and Jha, Debesh and Ghatwary, Noha and Realdon, Stefano and Cannizzaro, Renato and Salem, Osama E and Lamarque, Dominique and Daul, Christian and Riegler, Michael A and Anonsen, Kim V and others},
  journal={Scientific Data},
  volume={10},
  number={1},
  pages={75},
  year={2023},
  publisher={Nature Publishing Group UK London}
}

@article{smedsrud2021kvasir,
  title={Kvasir-Capsule, a video capsule endoscopy dataset},
  author={Smedsrud, Pia H and Thambawita, Vajira and Hicks, Steven A and Gjestang, Henrik and Nedrejord, Oda Olsen and N{\ae}ss, Espen and Borgli, Hanna and Jha, Debesh and Berstad, Tor Jan Derek and Eskeland, Sigrun L and others},
  journal={Scientific Data},
  volume={8},
  number={1},
  pages={142},
  year={2021},
  publisher={Nature Publishing Group UK London}
}

@article{boers2024foundation,
  title={Foundation models in gastrointestinal endoscopic AI: Impact of architecture, pre-training approach and data efficiency},
  author={Boers, Tim GW and Fockens, Kiki N and van der Putten, Joost A and Jaspers, Tim JM and Kusters, Carolus HJ and Jukema, Jelmer B and Jong, Martijn R and Struyvenberg, Maarten R and de Groof, Jeroen and Bergman, Jacques J and others},
  journal={Medical Image Analysis},
  volume={98},
  pages={103298},
  year={2024},
  publisher={Elsevier}
}

@inproceedings{bravo2025self,
  title={Self-Supervised Learning for Multi-Category Endoscopy Classification and Data Quality Evaluation Using Masked Autoencoders},
  author={Bravo, Diego and Ruano, Josu{\'e} and G{\'o}mez, Mart{\'\i}n and Gonz{\'a}lez, Fabio A and Romero, Eduardo},
  booktitle={2025 IEEE 22nd International Symposium on Biomedical Imaging (ISBI)},
  pages={1--5},
  year={2025},
  organization={IEEE}
}

@article{bravo2025gastrohun,
  title={Gastrohun an endoscopy dataset of complete systematic screening protocol for the stomach},
  author={Bravo, Diego and Frias, Juan and Vera, Felipe and Trejos, Juan and Mart{\'\i}nez, Carlos and G{\'o}mez, Mart{\'\i}n and Gonz{\'a}lez, Fabio and Romero, Eduardo},
  journal={Scientific Data},
  volume={12},
  number={1},
  pages={102},
  year={2025},
  publisher={Nature Publishing Group UK London}
}

@inproceedings{jha2023gastrovision,
  title={Gastrovision: A multi-class endoscopy image dataset for computer aided gastrointestinal disease detection},
  author={Jha, Debesh and Sharma, Vanshali and Dasu, Neethi and Tomar, Nikhil Kumar and Hicks, Steven and Bhuyan, Manas Kamal and Das, Pradip K and Riegler, Michael A and Halvorsen, P{\aa}l and Bagci, Ulas and others},
  booktitle={Workshop on machine learning for multimodal healthcare data},
  pages={125--140},
  year={2023},
  organization={Springer}
}

@inproceedings{DBLP:conf/iclr/DosovitskiyB0WZ21,
  author       = {Alexey Dosovitskiy and
                  Lucas Beyer and
                  Alexander Kolesnikov and
                  Dirk Weissenborn and
                  Xiaohua Zhai and
                  Thomas Unterthiner and
                  Mostafa Dehghani and
                  Matthias Minderer and
                  Georg Heigold and
                  Sylvain Gelly and
                  Jakob Uszkoreit and
                  Neil Houlsby},
  title        = {An Image is Worth 16x16 Words: Transformers for Image Recognition
                  at Scale},
  booktitle    = {9th International Conference on Learning Representations, {ICLR} 2021,
                  Virtual Event, Austria, May 3-7, 2021},
  year         = {2021}
}

@inproceedings{DBLP:conf/iclr/0097LL000NS23,
  author       = {Hao Zhang and
                  Feng Li and
                  Shilong Liu and
                  Lei Zhang and
                  Hang Su and
                  Jun Zhu and
                  Lionel M. Ni and
                  Heung{-}Yeung Shum},
  title        = {{DINO:} {DETR} with Improved DeNoising Anchor Boxes for End-to-End
                  Object Detection},
  booktitle    = {The Eleventh International Conference on Learning Representations,
                  {ICLR} 2023, Kigali, Rwanda, May 1-5, 2023},
  year         = {2023}
}

@inproceedings{he2022masked,
  title={Masked autoencoders are scalable vision learners},
  author={He, Kaiming and Chen, Xinlei and Xie, Saining and Li, Yanghao and Doll{\'a}r, Piotr and Girshick, Ross},
  booktitle={Proceedings of the IEEE/CVF conference on computer vision and pattern recognition},
  pages={16000--16009},
  year={2022}
}

@inproceedings{liu2022convnet,
  title={A convnet for the 2020s},
  author={Liu, Zhuang and Mao, Hanzi and Wu, Chao-Yuan and Feichtenhofer, Christoph and Darrell, Trevor and Xie, Saining},
  booktitle={Proceedings of the IEEE/CVF conference on computer vision and pattern recognition},
  pages={11976--11986},
  year={2022}
}

@inproceedings{huang2017densely,
  title={Densely connected convolutional networks},
  author={Huang, Gao and Liu, Zhuang and Van Der Maaten, Laurens and Weinberger, Kilian Q},
  booktitle={Proceedings of the IEEE conference on computer vision and pattern recognition},
  pages={4700--4708},
  year={2017}
}

@article{jia2020patient,
  title={A patient-similarity-based model for diagnostic prediction},
  author={Jia, Zheng and Zeng, Xian and Duan, Huilong and Lu, Xudong and Li, Haomin},
  journal={International journal of medical informatics},
  volume={135},
  pages={104073},
  year={2020},
  publisher={Elsevier}
}

@inproceedings{gruger2024enhancing,
  title={Enhancing Healthcare Decision-Making with Analogy-Based Reasoning},
  author={Gr{\"u}ger, Joscha and Kuhn, Martin and Amri, Karim and Bergmann, Ralph},
  booktitle={International Conference on Process Mining},
  pages={447--459},
  year={2024},
  organization={Springer}
}

@article{kim2020few,
  title={Few-shot visual reasoning with meta-analogical contrastive learning},
  author={Kim, Youngsung and Shin, Jinwoo and Yang, Eunho and Hwang, Sung Ju},
  journal={Advances in Neural Information Processing Systems},
  volume={33},
  pages={16846--16856},
  year={2020}
}

@article{doumas2022theory,
  title={A theory of relation learning and cross-domain generalization.},
  author={Doumas, Leonidas AA and Puebla, Guillermo and Martin, Andrea E and Hummel, John E},
  journal={Psychological review},
  volume={129},
  number={5},
  pages={999},
  year={2022},
  publisher={American Psychological Association}
}

@article{jin2024few,
  title={Few-shot learning with task adaptation for multi-category gastrointestinal endoscopy classification},
  author={Jin, Jun and Hu, Dasha and Pu, Wei and Luo, Yining and Feng, Xinyue},
  journal={Biomedical Signal Processing and Control},
  volume={95},
  pages={106387},
  year={2024},
  publisher={Elsevier}
}

@article{sheller2020federated,
  title={Federated learning in medicine: facilitating multi-institutional collaborations without sharing patient data},
  author={Sheller, Micah J and Edwards, Brandon and Reina, G Anthony and Martin, Jason and Pati, Sarthak and Kotrotsou, Aikaterini and Milchenko, Mikhail and Xu, Weilin and Marcus, Daniel and Colen, Rivka R and others},
  journal={Scientific reports},
  volume={10},
  number={1},
  pages={12598},
  year={2020},
  publisher={Nature Publishing Group UK London}
}

@inproceedings{sheller2018multi,
  title={Multi-institutional deep learning modeling without sharing patient data: A feasibility study on brain tumor segmentation},
  author={Sheller, Micah J and Reina, G Anthony and Edwards, Brandon and Martin, Jason and Bakas, Spyridon},
  booktitle={International MICCAI Brainlesion Workshop},
  pages={92--104},
  year={2018},
  organization={Springer}
}

@article{ngiam2019big,
  title={Big data and machine learning algorithms for health-care delivery},
  author={Ngiam, Kee Yuan and Khor, Wei},
  journal={The Lancet Oncology},
  volume={20},
  number={5},
  pages={e262--e273},
  year={2019},
  publisher={Elsevier}
}

@article{wahab2024federated,
  title={Federated deep learning for wireless capsule endoscopy analysis: Enabling collaboration across multiple data centers for robust learning of diverse pathologies},
  author={Wahab, Haroon and Mehmood, Irfan and Ugail, Hassan and Del Ser, Javier and Muhammad, Khan},
  journal={Future Generation Computer Systems},
  volume={152},
  pages={361--371},
  year={2024},
  publisher={Elsevier}
}

@article{li2025challenges,
  title={From challenges and pitfalls to recommendations and opportunities: Implementing federated learning in healthcare},
  author={Li, Ming and Xu, Pengcheng and Hu, Junjie and Tang, Zeyu and Yang, Guang},
  journal={Medical Image Analysis},
  pages={103497},
  year={2025},
  publisher={Elsevier}
}

@techreport{senthil2024benchmarking,
  title={Benchmarking and Boosting Localizers for Chest X-ray Images},
  author={Senthil Velan, Shivasakthi},
  year={2024},
  institution={Arizona State University}
}

@techreport{saravanan2024benchmarking,
  title={Benchmarking and Boosting of 3D Segmentation Models},
  author={Saravanan, Madhumitha},
  year={2024},
  institution={Arizona State University}
}

@inproceedings{islam2025foundation,
  title={Foundation X: integrating classification, localization, and segmentation through lock-release pretraining strategy for chest X-ray analysis},
  author={Islam, Nahid Ul and Ma, DongAo and Pang, Jiaxuan and Velan, Shivasakthi Senthil and Gotway, Michael and Liang, Jianming},
  booktitle={2025 IEEE/CVF Winter Conference on Applications of Computer Vision (WACV)},
  pages={3647--3656},
  year={2025},
  organization={IEEE}
}

@article{chen2023minigpt,
  title={Minigpt-v2: large language model as a unified interface for vision-language multi-task learning},
  author={Chen, Jun and Zhu, Deyao and Shen, Xiaoqian and Li, Xiang and Liu, Zechun and Zhang, Pengchuan and Krishnamoorthi, Raghuraman and Chandra, Vikas and Xiong, Yunyang and Elhoseiny, Mohamed},
  journal={arXiv preprint arXiv:2310.09478},
  year={2023}
}

@article{liu2023visual,
  title={Visual instruction tuning},
  author={Liu, Haotian and Li, Chunyuan and Wu, Qingyang and Lee, Yong Jae},
  journal={Advances in neural information processing systems},
  volume={36},
  pages={34892--34916},
  year={2023}
}

@inproceedings{liu2024improved,
  title={Improved baselines with visual instruction tuning},
  author={Liu, Haotian and Li, Chunyuan and Li, Yuheng and Lee, Yong Jae},
  booktitle={Proceedings of the IEEE/CVF conference on computer vision and pattern recognition},
  pages={26296--26306},
  year={2024}
}

@article{he2024efficient,
  title={Efficient multimodal learning from data-centric perspective},
  author={He, Muyang and Liu, Yexin and Wu, Boya and Yuan, Jianhao and Wang, Yueze and Huang, Tiejun and Zhao, Bo},
  journal={arXiv preprint arXiv:2402.11530},
  year={2024}
}

@article{li2024mini,
  title={Mini-gemini: Mining the potential of multi-modality vision language models},
  author={Li, Yanwei and Zhang, Yuechen and Wang, Chengyao and Zhong, Zhisheng and Chen, Yixin and Chu, Ruihang and Liu, Shaoteng and Jia, Jiaya},
  journal={arXiv preprint arXiv:2403.18814},
  year={2024}
}

@article{chu2023mobilevlm,
  title={Mobilevlm: A fast, reproducible and strong vision language assistant for mobile devices},
  author={Chu, Xiangxiang and Qiao, Limeng and Lin, Xinyang and Xu, Shuang and Yang, Yang and Hu, Yiming and Wei, Fei and Zhang, Xinyu and Zhang, Bo and Wei, Xiaolin and others},
  journal={arXiv preprint arXiv:2312.16886},
  volume={2},
  number={6},
  pages={7},
  year={2023}
}

@article{li2023llava,
  title={Llava-med: Training a large language-and-vision assistant for biomedicine in one day},
  author={Li, Chunyuan and Wong, Cliff and Zhang, Sheng and Usuyama, Naoto and Liu, Haotian and Yang, Jianwei and Naumann, Tristan and Poon, Hoifung and Gao, Jianfeng},
  journal={Advances in Neural Information Processing Systems},
  volume={36},
  pages={28541--28564},
  year={2023}
}

@article{mushtaq2026ai,
  title={AI-Assisted Double-Headed Capsule Endoscopy: Multicentre Prospective Diagnostic Accuracy Study Across Small Bowel Indications},
  author={Mushtaq, Kamran and Lim, Yun Jeong and Spada, Cristiano and Mussetto, Alessandro and Koulaouzidis, Anastasios and Kaung, Thake and Borrow, Dean-Martin and Casadei, Cesare and Patel, Praful and Rahman, Imdadur},
  journal={Diagnostics},
  volume={16},
  number={2},
  pages={239},
  year={2026},
  publisher={MDPI}
}

\end{document}